%% file: arxiv-tree-smc.tex
\pdfoutput=1        %
\documentclass[reqno,oneside,letterpaper]{amsart}
\usepackage{amsaddr} 
\usepackage[utf8]{inputenc} 

\usepackage{graphicx} %
\usepackage{subfigure} 
\usepackage[round]{natbib}
\usepackage{afterpage}
\usepackage{algorithm,algorithmic}
\usepackage[usenames,dvipsnames]{xcolor}
\usepackage[colorlinks,citecolor=blue,urlcolor=blue,linkcolor=RawSienna]{hyperref}

\usepackage{bm}
\usepackage{dsfont}
\usepackage{wrapfig}
\usepackage{amsmath,amssymb}
\usepackage{color}
\usepackage{soul}
\usepackage{wasysym}
\usepackage{grffile}
\usepackage{float}
\usepackage{rotating}	%

\newlength{\figwidth}
\newlength{\sfigwidth}
\input{macros.tex}

\def\wt{\tilde{w}}
\def\wtbar{\bar{\wt}}

\def\Pr{\mathbb{P}}

\def\given{\,|\,}
\def\defas{:=}
\def\Reals{\mathbb{R}}
\def\[#1\]{\begin{align}#1\end{align}}

\def\pendigits{\emph{pen-digits}}
\def\magic{\emph{magic-04}}

\def\uniformmp{\emph{empirical}}
\def\empirical{\emph{empirical}}
\def\posterior{\emph{optimal}}
\def\optimal{\emph{optimal}}
\def\prior{\emph{prior}}
\def\madelon{\emph{madelon}}
\newcommand{\TS}{\mathcal{T}}
\def\nodewise{\emph{node-wise}}
\def\layerwise{\emph{layer-wise}}
\def\nislands{I}%
\def\i{i}
\def\n{n}
\newcommand{\Cc}{m}
\newcommand{\cc}{{(m)}}
\newcommand{\Ccd}{{m'}}
\newcommand{\ccd}{{(m')}}
\newcommand{\nparts}{M}
\renewcommand\wt{w}

\usepackage{enumitem} %
\setenumerate{noitemsep} %
\setenumerate{itemindent=0pt,leftmargin=18pt}

\newcommand{\defn}[1]{{\bf #1}}
\renewcommand{\st}{\,:\,}

\renewcommand{\T}{\mathsf{T}}

\newcommand{\leaf}[1]{\partial #1}
\newcommand{\nonleaf}[1]{#1 \setminus \leaf {#1} }

\newcommand{\block}{B}

\newcommand{\Kr}{\mathbb{Q}}

\begin{document}

\title
[Top-down particle filtering for Bayesian decision trees]
{Top-down particle filtering\\for Bayesian decision trees}

\author{Balaji Lakshminarayanan}
\address{Gatsby Unit, CSML, University College London}

\author{Daniel M. Roy}
\address{Department of Engineering, University of Cambridge}

\author{Yee Whye Teh}
\address{Department of Statistics, University of Oxford}

\maketitle
\thispagestyle{empty}

\let\thefootnote\relax\footnote{This version is identical in content to ``Top-down particle filtering for Bayesian decision trees'' in Proceedings of the 30th International Conference on Machine Learning (ICML 2013), and differs only in typographic layout.}

\begin{abstract}
Decision tree learning is a popular approach for classification and regression in machine learning and statistics, and Bayesian formulations---which introduce a prior distribution over decision trees, and formulate learning as posterior inference given data---have been shown to produce competitive performance.  Unlike classic decision tree learning algorithms like ID3, C4.5 and CART, which work in a top-down manner, existing Bayesian algorithms produce an approximation to the posterior distribution by evolving a \emph{complete} tree (or collection thereof) iteratively via local Monte Carlo modifications to the structure of the tree, e.g., using Markov chain Monte Carlo (MCMC).  We present a sequential Monte Carlo (SMC) algorithm that instead works in a top-down manner, mimicking the behavior and speed of classic algorithms.  We demonstrate empirically that our approach delivers accuracy comparable to the most popular MCMC method, but operates more than an order of magnitude faster, and thus represents a better computation-accuracy tradeoff.
\end{abstract}

\section{Introduction}

Decision tree learning algorithms are widely used across statistics and machine learning, and often deliver near state-of-the-art performance despite their simplicity.  Decision trees represent predictive models from an input space, typically $\Reals^D$, to an output space of labels, and work by specifying 
a hierarchical partition of the input space into blocks.  Within each block of the input space, a simple model predicts labels.

In classical decision tree learning, a decision tree (or collection thereof) is learned in a greedy, top-down manner from the examples.
Examples of classical approaches that learn single trees include 
ID3 \citep{Quinlan:1986}, 
C4.5 \citep{quinlan1993c4} and 
CART \citep{breiman1984classification}, while methods that learn combinations of decisions trees include 
boosted decision trees \citep{friedman2001greedy},
Random Forests \citep{Breiman01},
and many others.

Bayesian decision tree methods, like those first proposed by 
\citet{Bun92},
\citet{chipman1998bayesian},
\citet{denison1998bayesian}, 
and 
\citet{chipman2000hierarchical}, 
and more recently revisited by 
\citet{wu2007bayesian}, 
\citet{taddy2011} and 
\citet{ag2012dynamic}, cast the problem of decision tree learning into the framework of Bayesian inference.  In particular, Bayesian approaches start by placing a prior distribution on the decision tree itself.  To complete the specification of the model, it is common to associate each leaf node with a parameter indexing a family of likelihoods, e.g., the means of Gaussians or Bernoullis.  The labels are then assumed to be conditionally independent draws from their respective likelihoods.
The Bayesian approach has a number of useful properties: e.g., the posterior distribution on the decision tree can be interpreted as reflecting residual uncertainty and can be used to produce point and interval estimates.  

On the other hand, exact posterior computation is typically infeasible and so existing approaches use approximate methods such as Markov chain Monte Carlo (MCMC) in the batch setting.
Roughly speaking, these algorithms iteratively improve a complete decision tree by making a long sequence of random, local modifications, each biased towards tree structures with higher posterior probability.  These algorithms stand in marked contrast with classical decision tree learning algorithms like ID3 and C4.5, which rapidly build a decision tree for a data set in a top-down greedy fashion guided by heuristics.  Given the success of these methods, one might ask whether they could be adapted to work in the Bayesian framework.

In this article, we present such an adaptation, proposing a sequential
Monte Carlo (SMC) method for approximate inference in Bayesian decision trees that works by sampling a collection of trees in a top-down manner like ID3 and C4.5.  Unlike classical methods, there is no pruning stage after the top-down learning stage to prevent over-fitting, as the prior combines with the likelihood to automatically cut short the growth of the trees, and resampling focuses attention on those trees that better fit the data.  In the end, the algorithm produces a collection of sampled trees that approximate the posterior distribution.   While both existing MCMC algorithms and our novel SMC algorithm produce approximations to the posterior that are exact in the limit, 
we show empirically that our algorithms run more than an order of magnitude faster than existing methods while delivering the same predictive performance.

The article is organized as follows: we begin by describing the Bayesian decision tree model precisely in Section~\ref{sec:model}, and then describe the SMC algorithm in detail in Section~\ref{sec:smc}.  Through a series of empirical tests, we demonstrate in Section~\ref{sec:experiments} that this approach is fast and produces good approximations.  We conclude in Section~\ref{sec:discussion} with a discussion comparing this approach with existing ones in the Bayesian setting, and point towards future avenues.

\section{Model and notation}\label{sec:model}

\newcommand{\LatentTree}{\mathcal T}
In this section, we present the decision tree model for the distribution of the labels $Y=\{y_\n\}_{\n=1}^N$ corresponding to input vectors $\bX=\{\bx_\n\}_{\n=1}^N$, $\bx_\n\in \Reals^D$.  The assumption is that the probabilistic mapping from input vectors to their labels is mediated by a latent decision tree $\LatentTree$ that serves to partition the input space into axis-aligned blocks.  Each block is then associated with a parameter that determines the distribution of the labels of the input vectors falling in that block.

\begin{figure}[t!]
\centering
\vspace{6mm}
\includegraphics[height=3.5cm]{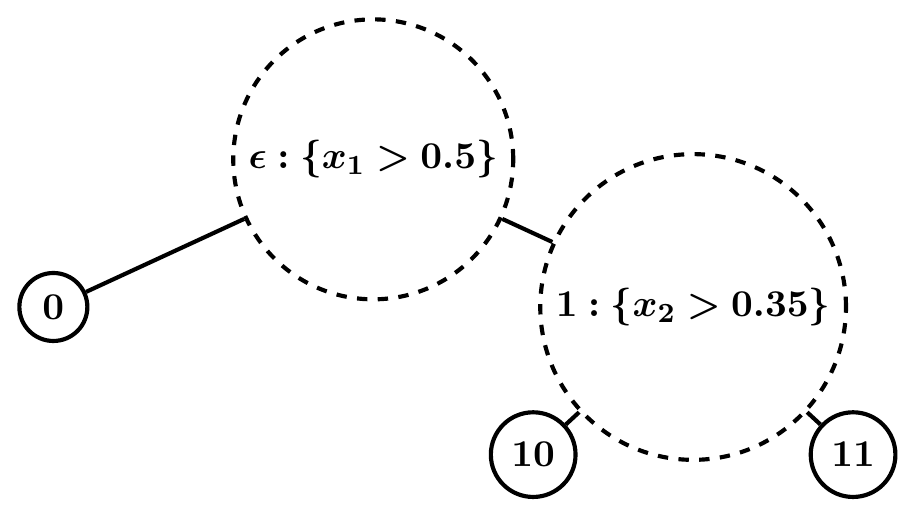}
\includegraphics[height=3.5cm]{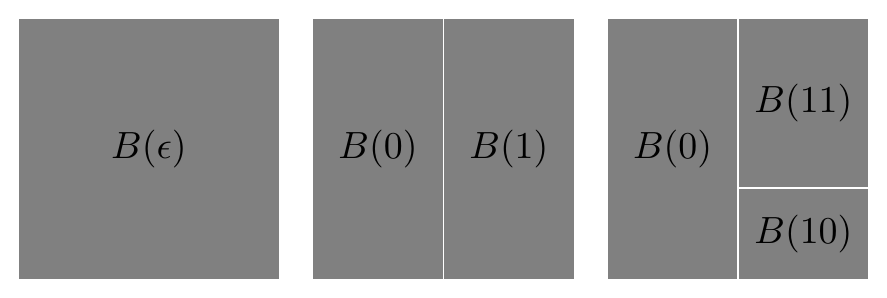}
\includegraphics[height=3.5cm]{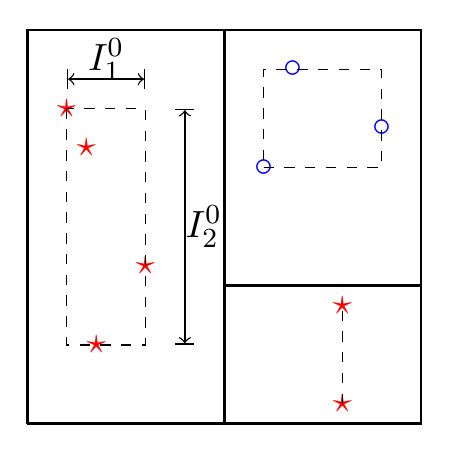}
\caption{
A decision tree $\LatentTree=(\T,\kappa,\tau)$ represents a hierarchical partitioning of a space.  Here, the space is the unit square and the tree $\T$ contains the nodes $\{\epsilon, 0, 1, 10, 11\}$.  The root node $\epsilon$ represents the whole space $B(\epsilon) = \Reals^D$, while its two children $0$ and $1$, represent the two halves of the cut $(\kappa(\epsilon),\tau(\epsilon)) = (1,0.5)$, where $\kappa(\epsilon)=1$ represents the dimension of the cut, and $\tau(\epsilon) = 0.5$ represents the location of the cut along that dimension.  (The origin is at the bottom left of each figure, and the $x$-axis is dimension 1. The red stars and blue circles represent observed data points.) The second cut, $(\kappa(1),\tau(1))=(2,0.35)$, splits the block $B(1)$ into the two halves $B(11)$ and $B(10)$.  When defining the prior over decision trees given by \citet{chipman1998bayesian}, it will be necessary to refer to the ``extent'' of the data in a block.  E.g., $I_1^0$ and $I_2^0$ are the extent of the data in dimensions 1 and 2, respectively, in block $B(0)$.  For each node $\pp$, the set $D^\pp$  contains those dimensions with non-trivial extent.  Here, $D^0 = \{1,2\}$, but $D^{10} = \{2\}$, because there is no variation in dimension 1. \vspace{10mm}
}
\label{mainfigure}
\end{figure}

A rooted, strictly binary tree $\T$ is a finite tree with a single root, denoted by the empty string $\epsilon$, where each internal node $\pp$ except the root has exactly two children, called the left child $\pp0$ and the right child $\pp1$.  Denote the leaves of $\T$ (those nodes without children) by $\leaf\T$.  Each node of the tree $\pp\in\T$  is associated with a block $B(\pp)\subset\Reals^D$ of the input space as follows:  At the root we have $B(\epsilon)=\Reals^D$, while each internal node $\pp\in\nonleaf\T$ ``cuts'' its block into two halves, with $\kappa(\pp)\in\{1,\dotsc,D\}$ denoting the dimension of the cut, and $\tau(\pp)$ denoting the location of the cut, so that
\begin{align}
B(\pp0) &= B(\pp)\cap\{\bz \in\Reals^D: z_{\kappa(\pp)}\le \tau(\pp)\} \text{ and }\nonumber \\
B(\pp1) &= B(\pp)\cap\{\bz \in\Reals^D: z_{\kappa(\pp)}> \tau(\pp)\}.
\end{align}
We call the tuple $\LatentTree=(\T,\kappa,\tau)$ the decision tree.  (See Figure~\ref{mainfigure} for more intuition on the representation and notation of decision trees.)   Note that the blocks associated with the leaves of the tree partition $\Reals^D$.
It will be convenient to write $N(\pp)$ for the set of data point indices $\n \in \{1,\dots,N\}$ such that $\bx_\n \in \block (\pp)$.
For every subset $A \subseteq \{1,\dotsc,N\}$, let $Y_A \defas \{y_\n \st \n \in A\}$ and similarly for $\bX_A$, 
so that $\bX_{N(\pp)}$ are the input vectors in block $\block(\pp)$ and $Y_{N(\pp)}$ are their labels.   Note that both $B(\pp)$ and $N(\pp)$ depend on $\LatentTree$, although we have chosen to elide this dependence for notational simplicity.

Conditioned on the examples $\bX$, we assume that 
the joint density $f(Y,\LatentTree\given \bX)$ of the labels $Y$ and the latent decision tree $\LatentTree$ factorizes as follows:
\[
f(Y,\LatentTree \given \bX) 
&= h(\LatentTree \given \bX) \, g(Y \given \LatentTree, \bX) \nonumber \\
&\textstyle \label{likeeqn}
= h(\LatentTree \given \bX) \, \prod_{\pp \in \leaf \T} \ell( Y_{N(\pp)}|\bX_{N(\pp)} )
\]
where $\ell$ denotes a likelihood, defined below.

In this paper, we focus on the case of categorical labels taking values in the set $\{1,\dotsc,K\}$.  It is natural to take $\ell$ to be the Dirichlet-Multinomial likelihood, corresponding to the data being conditionally i.i.d.\ draws from a multinomial distribution on $\{1,\dotsc,K\}$ with a Dirichlet prior.  In particular,
\[
\ell (Y_{N(\pp)}|\bX_{N(\pp)}) = 
  \frac{\Gamma(\alpha)}{\Gamma(\frac{\alpha}{K})^K}
                             \frac{\prod_{k=1}^K \Gamma(m_{pk}+\frac{\alpha}{K})}
                                            {\Gamma(\sum_{k=1}^K m_{pk} + \alpha)}, \label{eq:loglik}
\]
where $m_{pk}$ denotes the number of labels $y_\n=k$ among those $\n\in N(\pp)$ and
$\alpha$ is the concentration parameter of the symmetric Dirichlet prior.  Generalisations to other likelihood functions based on conjugate pairs of exponential families are straightforward.

The final piece of the model is the prior density $h(\LatentTree|\bX)$
over decision trees.  In order to make straightforward comparisons with existing algorithms, we adopt the model proposed by 
\citet{chipman1998bayesian}.  In this model, the prior distribution of the latent tree is defined \emph{conditionally} on the given input vectors $\bX$ (see Section~\ref{sec:discussion} for a discussion of this dependence on $\bX$ and its effect on the exchangeability of the labels).
Informally, the tree is grown starting at the root, and each new node either splits and grows two children (turning the node into an internal node) or stops (leaving it a leaf) stochastically.  

We now describe the generative process more precisely in terms of a Markov chain capturing the construction of a decision tree in stages, beginning with the trivial tree $\T_0=\{\epsilon\}$ containing only the root node.  At each stage $\i$, $\T_\i$ is produced from $\T_{\i-1}$ by choosing one leaf in $\T_{\i-1}$ and either \emph{growing} two children nodes or \emph{stopping} the leaf.  Once stopped, a leaf is ineligible for future growth.
The identity of the chosen leaf is \emph{deterministic}, while the choice to grow or stop is \emph{stochastic}.  The process proceeds until all leaves are stopped, and so
each node is considered for expansion exactly once throughout the process.
This will be seen to give rise to a finite sequence of decision trees $\LatentTree_\i = (\T_\i,\kappa_\i,\tau_\i)$ once we define the associated cut functions $\kappa_\i$ and $\tau_\i$.
We will use this Markov chain in Section~\ref{sec:smc} as scaffolding for a sequential Monte Carlo algorithm. A similar approach was employed by \citet{taddy2011} in the setting of online Bayesian decision trees.  There are similarities also with the \emph{bottom-up} SMC algorithms by \citet{TehDauRoy2008} and \citet{Bou12}.

We next describe the rule for stopping or growing nodes, and the distribution of cuts.
Let $\pp$ be the node chosen at some stage of the generative process.  If the input vectors $\bX_{N(\pp)}$ are all identical, then the node stops and becomes a leaf.  (Chipman et al.\ chose this rule because no choice of cut to the block $B(\pp)$ would result in both children containing at least one input vector.)
Otherwise, let $D^\pp$ be the set of dimensions along which $\bX_{N(\pp)}$ varies, and let $I^\pp_d=[\min_{\n \in N(\pp)} x_{\n d},\max_{\n\in N(\pp)} x_{\n d}]$ be the range of the input vectors along dimension $d\in D^\pp$.  (See last subfigure of Figure~\ref{mainfigure}.)
Under the Chipman et al.\ model, the probability that node $\pp$ is split is 
\[\label{eq:split}
\frac {\alpha_s} { (1+|\pp|)^{\beta_s}}\ , \qquad \alpha_s \in (0,1),\ \beta_s\in [0,\infty),
\]  
where $|\pp|$ is the depth of the node, and $\alpha_s$ and $\beta_s$ are parameters governing the shape of the resulting tree.  For larger $\alpha_s$ and smaller $\beta_s$ the typical trees are larger, while the deeper $\pp$ is in the tree the less likely it will be cut. If $\pp$ is cut, the dimension $\kappa(\pp)$ and then location $\tau(\pp)$ of the cut are sampled uniformly from $D^\pp$ and $I^\pp_{\kappa(\pp)}$, respectively.  Note that the choice for the support of the distribution over cut dimensions and locations are such that both children of $\pp$ will, with probability one, contain  at least one input vector.  Finally, the choices of whether to grow or stop, as well the cut dimensions and locations, are conditionally independent across different subtrees. 

To complete the generative model, we define $\T = \T_\eta$, $\kappa = \kappa_\eta$ and $\tau = \tau_\eta$, where $\eta$ is the first stage such that all nodes are stopped. We note that $\eta < 2N$ with probability one because each cut of a node $\pp$ produces a non-trivial partition of the data in the block, and a node with one data point will be stopped instead of cut. 
The conditional density of the decision tree $\LatentTree=(\T,\kappa,\tau)$ can now be expressed as
\[
h(\T,\kappa,\tau|\bX) 
 &= 
 \prod_{\pp \in \leaf \T} 
 \Bigl(1-\frac{\alpha_s}{(1+|\pp|)^{\beta_s}}\Bigr)^{ \indicator(|D^\pp|>0)} 
\ 
  \prod_{\pp \in \nonleaf \T } 
  \frac{\alpha_s}{(1+|\pp|)^{\beta_s}}
  \frac1{|D^\pp|} \frac1{|I^\pp_{\kappa(\pp)}|} \,.
\]
Note that the prior distribution of $\LatentTree$ does not depend on the deterministic rule for choosing a leaf at each stage.  However this choice will have an effect on the bias/variance of the corresponding SMC algorithm.  

\section{Sequential Monte Carlo (SMC) for Bayesian decision trees}\label{sec:smc}

In this section we describe an SMC algorithm for approximating the posterior distribution over the decision tree $(\T,\kappa,\tau)$ given the labeled training data $(\bX, Y)$. (We refer the reader to \citep{cappe2007overview} for an excellent overview of SMC techniques.)
The approach we will take is to perform particle filtering following the sequential description of the prior.  
In particular, at stage $\i$, the particles approximate a modified posterior distribution where the prior on $(\T,\kappa,\tau)$ is replaced by the distribution of $(\T_\i,\kappa_\i,\tau_\i)$, i.e., the process truncated at stage $\i$.

Let $E_\i$ denote the set of unstopped leaves at stage $i$, all of which are eligible for expansion.
An important freedom we have in our SMC algorithm is the choice of which candidate leaf (or set $C_i \subseteq E_i$ of candidate leaves) to consider expanding.
In order to avoid ``multipath'' issues \citep[][\S3.5]{MR2278333} which lead to high variance, 
we fix a \emph{deterministic} rule for choosing $C_\i \subseteq E_i$.  (Multiple candidates are expanded or stopped in turn, independently.)
This rule can be a function of $(\bX, Y)$ and the state of the current particle, as the correctness of resulting approximation is unaffected.
We evaluate two choices in experiments: first, the rule $C_\i = E_\i$ where we consider expanding all eligible nodes; and second, the rule where $C_\i$ contains a single node chosen in a breadth-first (i.e., oldest first) manner from $E_\i$.

We may now define the sequence $(\Pr^Y_\i)$ of \defn{target distributions}.  Recall the sequential process defined in Section~\ref{sec:model}.  If the generative process for the decision tree has not completed by stage $\i$, the process has generated $(\T_\i,\kappa_\i, \tau_\i)$ along with $E_\i$, capturing which leaves in $\T_\i$ have been considered for expansion in previous stages already and which have not.  Let $\TS_\i=(\T_\i,\kappa_\i,\tau_\i,E_\i)$ be the variables generated on stage $\i$, and write $\Pr$ for the prior distribution on the sequence $(\TS_\i)$.  We construct the target distribution $\Pr^Y_\i$ as follows: Given $\TS_\i$, we generate labels $Y'$ with likelihood $g(Y' |\LatentTree_\i,\bX)$, i.e., as if $(\T_\i,\kappa_\i,\tau_\i)$ were the complete decision tree.  We then define $\Pr^Y_\i$ to be the conditional distribution of $\TS_\i$ given $Y'=Y$.  That is, $\Pr^Y_\i$ is the posterior with a truncated prior.

In order to complete the description of our SMC method, we must define 
\defn{proposal kernels} $(\Kr_\i)$ that sample approximations for the $\i$th stage given values for the $(\i-1)$th  stage.
As with our choice of $C_\i$, we have quite a bit of freedom.  In particular, the proposals can depend on the training data $(\bX, Y)$.  
An obvious choice is to take $\Kr_\i$ to be the conditional distribution of $\TS_\i$ given $\TS_{\i-1}$ under the prior, i.e., setting $\Kr_\i(\TS_\i\given \TS_{\i-1}) = \Pr(\TS_\i \given \TS_{\i-1})$.
Informally, this choice would lead us to propose extensions to trees at each stage of the algorithm by sampling from the prior, so we will refer to this as the \defn{prior proposal} kernel (aka the 
Bayesian bootstrap filter \citep{gordon:107}).  

We consider two additional proposal kernels:  
The first,
\[
\Kr_\i(\TS_\i\given \TS_{\i-1}) = \Pr^Y_\i(\TS_\i \given \TS_{\i-1}),
\]
is called the \defn{(one-step) optimal proposal} kernel because it would be the optimal kernel assuming that the $\i$th stage were the final stage.  We return to discuss this kernel in Section~\ref{sec:onestep}. 
The second alternative, which we will refer to as the \defn{empirical proposal} kernel, is a small modification to the \prior\ proposal, differing only in the choice of the split point $\tau$.  Recall that, in the prior,
$\tau_\i(\pp)$ is chosen uniformly from the interval $I^\pp_{\kappa_\i(\pp)}$.  This ignores the empirical distribution given by the input data $\bX_{N(\pp)}$ in the partition. We can account for this by first choosing, uniformly at random, a pair of adjacent data points along feature dimension $\kappa_\i(\pp)$, and then sampling a cut $\tau_\i(\pp)$ uniformly from the interval between these two data points.

The pseudocode for our proposed SMC algorithm is given in Algorithm \ref{alg:smc} in Appendix \ref{sec:alg:smc}.  Note that the SMC framework only requires us to compute the density of $\TS_\i$ under the target distribution up to a normalization constant.
In fact, the SMC algorithm produces an estimate of the normalization constant, which, at the end of the algorithm, is equal %
to the marginal probability of the labels $Y$ given $\bX$, with the latent decision tree $\LatentTree$ marginalized out. %
In general, the joint density of a Markov chain can be hard to compute, but because the set of nodes $C_\i$ considered at each stage is a deterministic function of $\TS_\i$, 
the path $(\TS_0,\TS_1,\dotsc,\TS_{\i-1})$ taken is a deterministic function of $\TS_\i$.  As a result, the joint density is simply a product of probabilities for each stage.
The same property holds for the proposal kernels defined above because they use the same candidate set $C_\i$, and have the same support as $\Pr$.
These properties justify the equations in Algorithm \ref{alg:smc}.

\subsection{The one-step optimal proposal kernel}\label{sec:onestep}

In this section we revisit the definition of the one-step \optimal\ proposal kernel.  
While the \prior\ and \empirical\ proposal kernels are relatively straightforward, the one-step \optimal\ proposal kernel is defined in terms of an additional conditioning on the labels $Y$, which we now study in greater detail.

Recall that the one-step \optimal\ proposal kernel $\Kr_\i$ is given by 
$\Kr_\i(\TS_\i\given \TS_{\i-1}) = \Pr^Y_\i(\TS_\i \given \TS_{\i-1})$.  
To begin, we note that, conditionally on $\TS_{\i-1}$ and $Y$, 
the subtrees rooted at each node $\pp \in C_{\i-1}$ are independent.
This follows from 
the fact that the likelihood of $Y$ given $\TS_\i$ factorizes over the leaves.
Thus, the proposal's probability density is
\[%
\Kr_\i(\TS_\i | \TS_{\i-1})
= \prod_{\pp \in C_{\i-1}} Q_\i(\rho_{\i,\pp},\kappa_\i(\pp),\tau_\i(\pp)),
\]
where $Q_\i$ is the probability density of the cuts at node $\pp$ under $\Kr_\i$, and $\rho_{\i,\pp}$ denotes whether the node was split or not.
On the event we split a node $\pp \in C_{\i-1}$, if we condition further on $\kappa_\i(\pp)$ and $\rho_{\i,\pp}$, we note that the conditional likelihood of $Y_{N(\pp)}$,
when viewed as a function of the split $\tau_\i(\pp)$, is piecewise constant, and in particular, only changes when the split crosses an example.  It follows that we can sample from this proposal by first considering the discrete choice of an interval, and then sampling uniformly at random from within the interval, as with the \uniformmp\ proposal.  
Some algebra shows that
\[ %
Q_\i(\rho_{\i,\pp}=\text{stop}) 
\propto \
\Bigl(1- \frac{\alpha_s}{(1+|\pp|)^{\beta_s}}\Bigr) \, \ell (Y_{N(\pp)}|\bX_{N(\pp)}) \,, \nn
\]
and
\[
Q_\i(\rho_{\i,\pp}=\text{split},\kappa_\i(\pp),\tau_\i(\pp))
&\propto \frac{\alpha_s}{(1+|\pp|)^{\beta_s}} \frac 1 {|D^\pp|}  \frac 1 {|I^\pp_{\kappa_\i(\pp)}|} 
\prod_{j = 0,1} \ell (Y_{N(\pp j)}|\bX_{N(\pp j)}) .\nn
\]

\subsection{Computational complexity}\label{sec:complexity}%
Let $U_d$ denote the number of unique values in dimension $d$,  $N_\pp$ denote the number of training data points at node $\pp$ and $\eta^\cc$ denote the number of nodes in particle $\Cc$. For all the SMC algorithms, the space complexity is $\O(\nparts N) + \O(\sum_d U_d) + \O(\sum_\Cc\eta^\cc)$. The time complexity is $\O(D N\log N) + \nparts \sum_\pp \O(2DN_\pp + N_\pp) $ 
 for \prior\ and \empirical\ proposals 
and $\nparts \sum_\pp\bigl(D \O( N_\pp \log N_\pp 
) + N_\pp\bigr)$  for the \optimal\ proposal.  The \optimal\ proposal typically requires higher computational cost per particle, but fewer number of particles than the \prior\ and \empirical\ proposals.

\section{Experiments}\label{sec:experiments}
In this section, we experimentally evaluate the design choices of the SMC algorithm (proposal, expansion strategy, number of particles and ``islands'') on real world datasets. In addition, we compare the performance of SMC to the most popular MCMC method for Bayesian decision tree learning \citep{chipman1998bayesian}, as well as CART, a popular (non-Bayesian) tree induction algorithm.
We evaluate all the algorithms on the following datasets from the UCI ML repository \citep{Asuncion+Newman:2007}: 
\begin{itemize}[topsep=1pt,parsep=1pt,itemsep=1pt]{
\item MAGIC gamma telescope data 2004 (\magic): $N=19020$, $D=10$, $K=2$. 
\item Pen-based recognition of handwritten digits (\pendigits): $N=10992$, $D=16$, $K=10$.
}\end{itemize}
Previous work has focused mainly on small datasets (e.g., the Wisconsin breast cancer database used by \citet{chipman1998bayesian} has $683$ data points). We chose the above datasets to illustrate the scalability of our approach. For the \pendigits\ dataset, we used the predefined training/test splits, while for the other datasets, we split the datasets randomly into a training set  and a test set  containing approximately 70\% and 30\% of the data points respectively. 

We implemented our scripts in Python and applied similar software optimization techniques to SMC and MCMC scripts.\footnote{The scripts can be downloaded from the authors' webpages.}
 Our experiments were run on a cluster with machines of similar processing power.

\subsection{Design choices in the SMC algorithm}\label{sec:smc:design}
In these set of experiments, we fix the hyperparameters to $\alpha=5.0, \alpha_s=0.95, \beta_s=0.5$ and compare the predictive performance of different configurations of  the SMC algorithm for this fixed model. Under the prior, these values of  $\alpha_s, \beta_s$ produce trees whose mean depth and number of nodes are $5.1$ and $18.5$, respectively. Given $\nparts$ particles, we use an effective sample size (ESS) threshold of $\nparts/10$ and set the maximum number of stages to 5000 (although the algorithms never reached this number).%
\subsubsection{Proposal choice and node expansion}
We consider the SMC algorithm proposed in Section $\ref{sec:smc}$ under two proposals: \posterior\ and \prior. (The  \uniformmp\ proposal performed similar to the \prior\ proposal and hence we do not report those results here.) 
We consider two strategies for choosing $C_\i$, i.e., the list of nodes considered for expansion at stage $\i$: (i) \emph{node-wise expansion}, where a single node is considered for expansion per stage (i.e., $C_\i$ is a singleton chosen deterministically from eligible nodes $E_\i$), and 
 (ii) \emph{layer-wise expansion}, where all nodes at a particular depth are considered for expansion simultaneously (i.e., $C_\i=E_\i$). 
For \nodewise\ expansion, we evaluate two strategies for selecting the node deterministically from $C_\i$: (i) breadth-first priority, where the oldest node is picked first, and (ii) marginal-likelihood based priority, where we expand the node with the lowest marginal likelihood. Both of these priority schemes performed similarly; hence we report only the results for breadth-first priority. We use multinomial resampling in our experiments. We also evaluated systematic resampling \citep{douc2005comparison} but found that the performance was not  significantly different.

\begin{figure}
        \centering
        \begin{subfigure} 
                \centering
                \includegraphics[width=\figwidth]{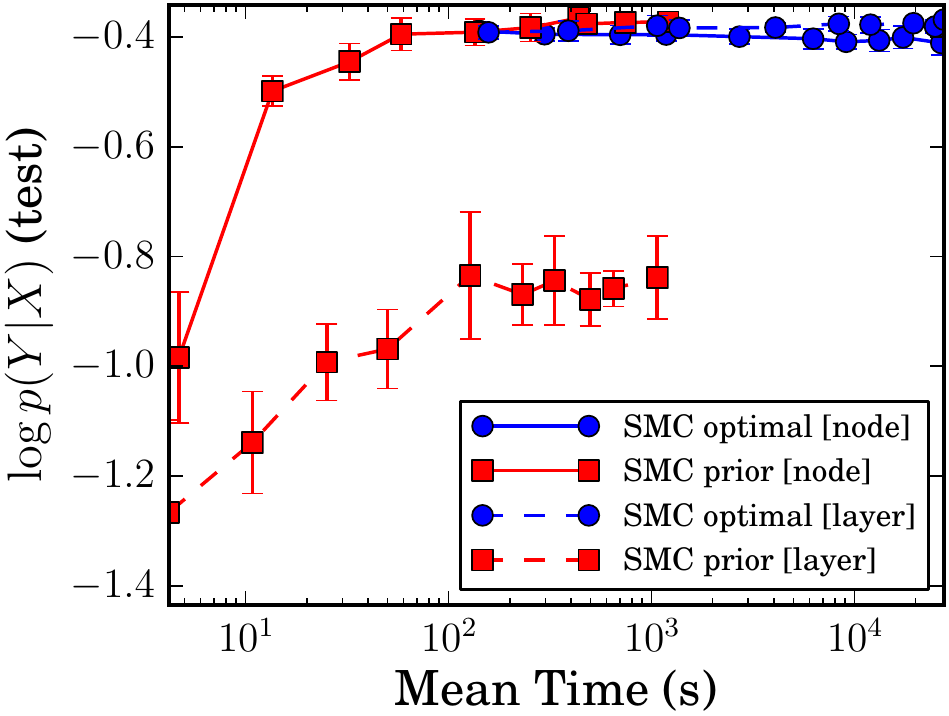}
        \end{subfigure}  \begin{subfigure} 
                \centering
                \includegraphics[width=\figwidth]{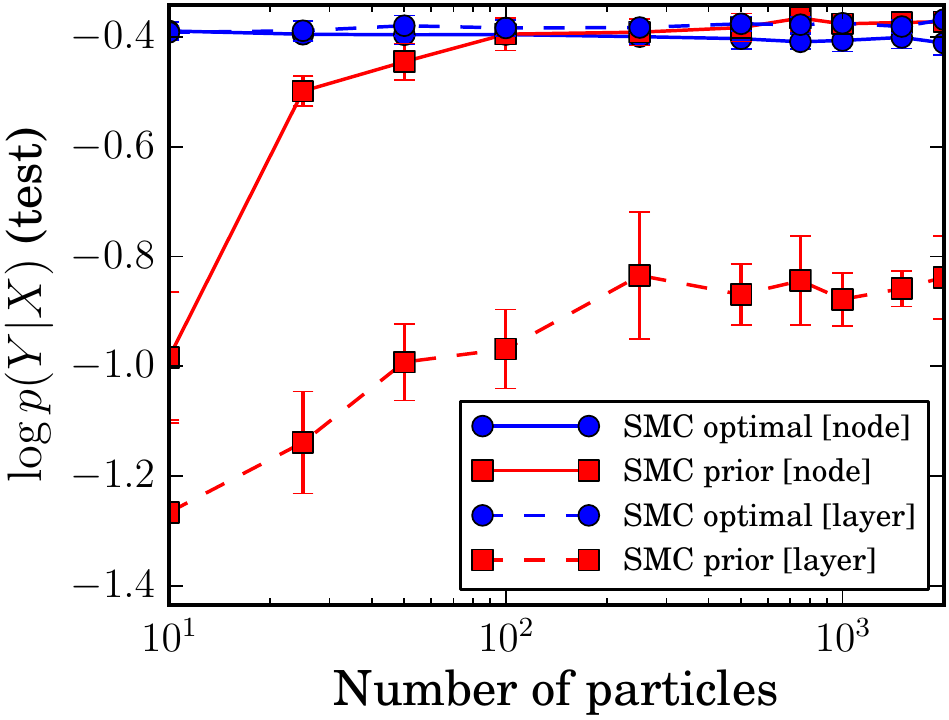}
        \end{subfigure}
        ~
                \begin{subfigure} 
                \centering
                \includegraphics[width=\figwidth]{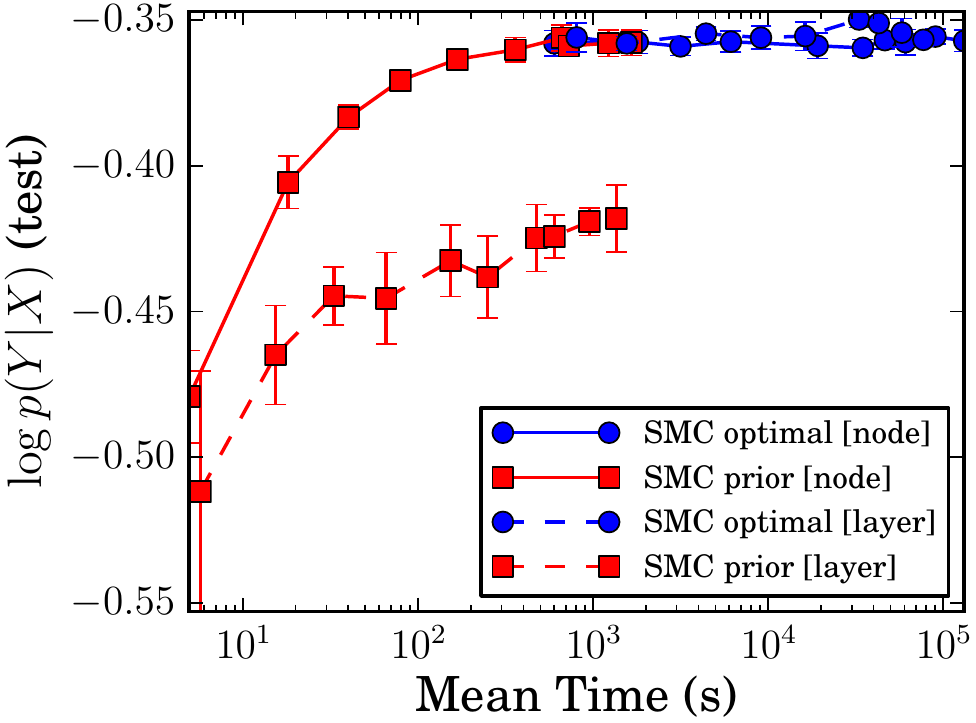}
        \end{subfigure}
        \begin{subfigure} 
                \centering
                \includegraphics[width=\figwidth]{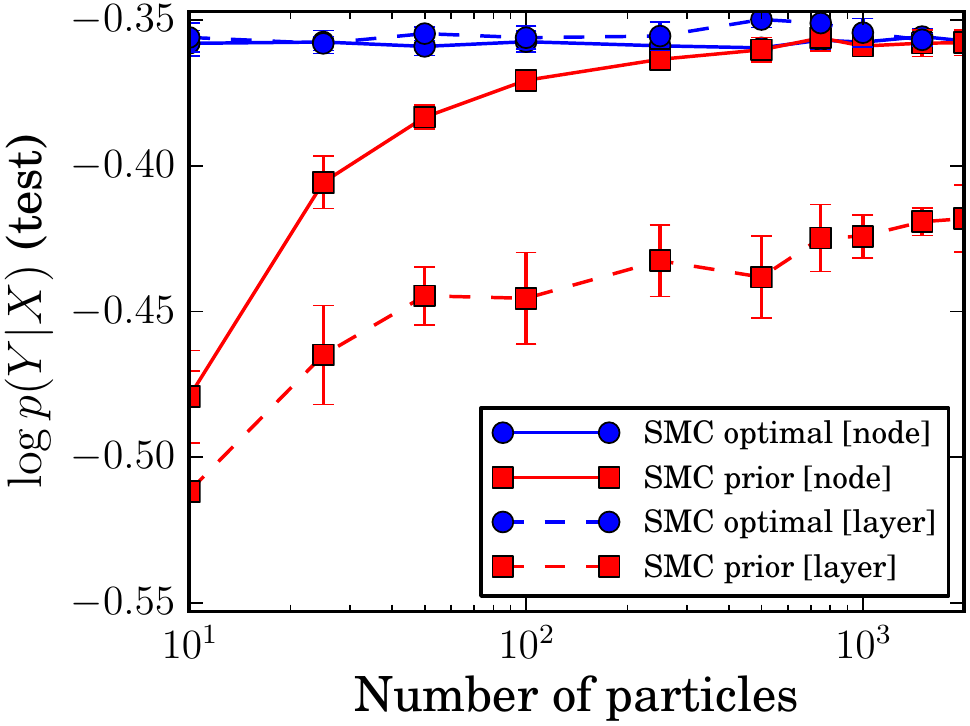}
        \end{subfigure}
        \caption{%
    Results on \pendigits\ (top), and \magic\ (bottom). Left column plots test $\log p(y|\bx)$ vs runtime, while right column plots test $\log p(y|\bx)$ vs number of particles. The blue circles and red squares represent \posterior\ and \prior\ proposals respectively. The solid and dashed lines represent \nodewise\ and \layerwise\ proposals respectively. }
        \label{fig:logprobtest:time:p}
\end{figure}

We report the log predictive probability on test data as a function of runtime and of the number of particles (similar trends are observed for test accuracy; see Appendix~\ref{sec:expt:smc:acctest}).  
The times reported do not account for prediction time. We average the numbers over 10 random initializations and report standard deviations. The results are shown in Figure \ref{fig:logprobtest:time:p}. In summary, we observe the following: 
\begin{enumerate}[topsep=1pt,parsep=1pt,itemsep=1pt]{ %
\item \nodewise\ expansion outperforms \layerwise\ expansion for \prior\ proposal. The \prior\ proposal does not account for likelihood; one could think of the resampling steps as `correction steps' for the sub-optimal decisions sampled from the \prior\ proposal. Because  \nodewise\ expansion can potentially resample at every stage, it can correct individual bad decisions immediately, whereas \layerwise\ expansion cannot.
  In particular, we have observed that \layerwise\ expansion tends to produce shallower trees compared to \nodewise\ expansion, leading to poorer performance. This phenomenon can be explained as follows: as the depth of the node increases, the prior probability of stopping increases whereas the posterior probability of stopping might be quite low. In \nodewise\ expansion, the resampling step can potentially retain the particles where the node has not been stopped. However, in \layerwise\
  expansion, too many nodes might have stopped prematurely and the resampling step cannot `correct' all these bad decisions easily (i.e., it would require many more particles to sample trees where all the nodes in a layer have not been stopped). Another interesting observation is that  \layerwise\ expansion exhibits higher variance: this can be explained by the fact that \layerwise\ expansion samples a greater number of random variables (on average) than \nodewise\ before resampling, and so suffers for the same reason that importance sampling can suffer from high variance. Note that both expansion strategies perform
  similarly for the \posterior\ proposal due to the fact that the proposal %
  accounts for the likelihood
  and resampling does not affect the results significantly. Due to its superior performance, we consider only \nodewise\ expansion in the rest of the paper.

\item The plots on the right side of Figure \ref{fig:logprobtest:time:p} suggest that the \posterior\ proposal requires fewer particles than the \prior\ proposal (as expected). 
However, the per-stage cost of \posterior\ proposal is much higher than the \prior, leading to significant increase in the overall runtime (see Section \ref{sec:complexity} for a related discussion). Hence, the \emph{prior} proposal offers a better predictive performance vs computation time tradeoff than the \emph{optimal} proposal.  
\item The performance of \posterior\ proposal saturates very quickly and is near-optimal even when the number of particles is small ($\nparts=10$). 
}\end{enumerate}

 \subsubsection{Effect of irrelevant features}\label{sec:irrelevant}
In the next experiment, we test the effect of irrelevant features on the performance of the various proposals. We use the \madelon\ dataset\footnote{\url{http://archive.ics.uci.edu/ml/datasets/Madelon}} for this experiment, in which the data points belong to one of 2 classes and lie in a 500-dimensional space, out of which only 20 dimensions are deemed relevant. %
The training dataset contains 2000 data points and the test dataset contains 600 data points.  We use the validation dataset in the UCI ML repository as our test set because labels are not available for the test dataset.

\begin{figure}
        \centering
        \begin{subfigure} 
                \centering
                \includegraphics[width=\sfigwidth]{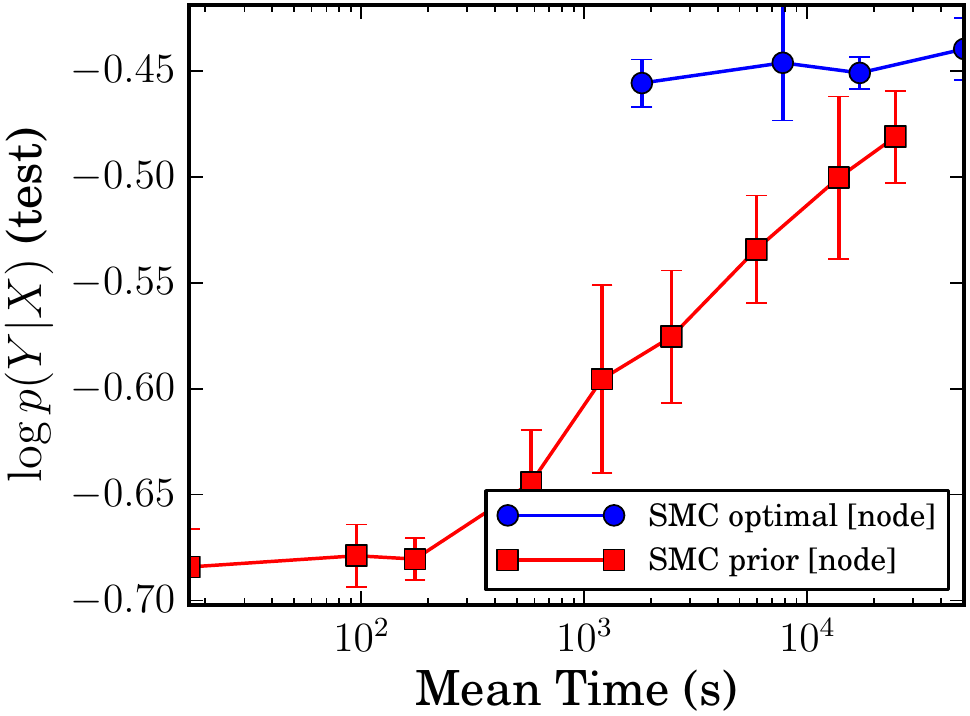}
        \end{subfigure}  \begin{subfigure} 
                \centering
                \includegraphics[width=\sfigwidth]{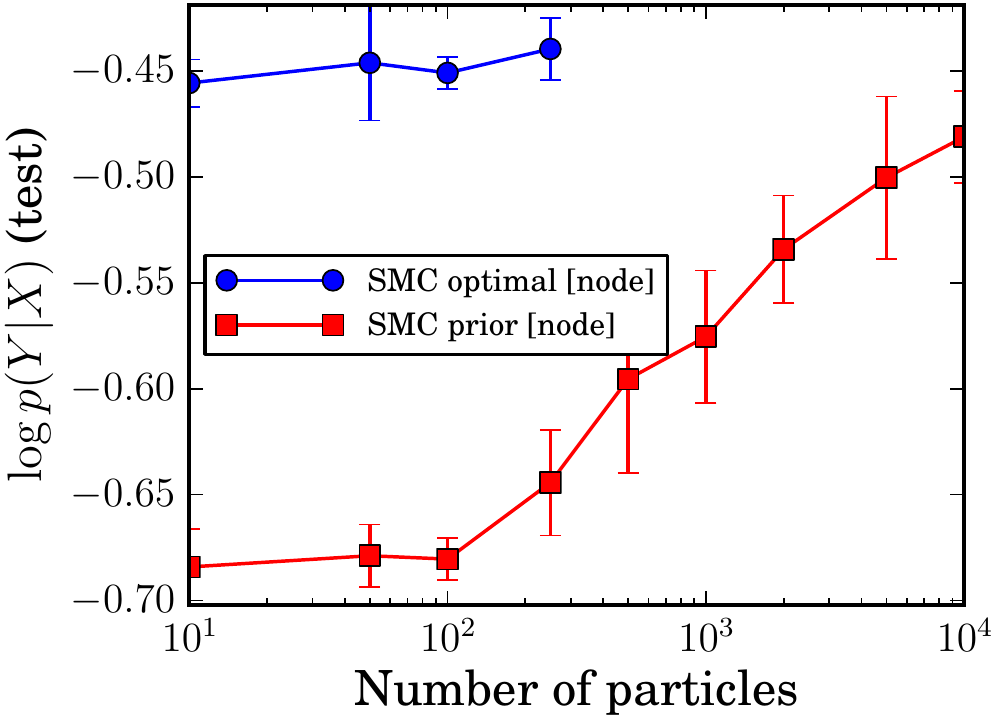}
        \end{subfigure}
        ~
                \begin{subfigure} 
                \centering
                \includegraphics[width=\sfigwidth]{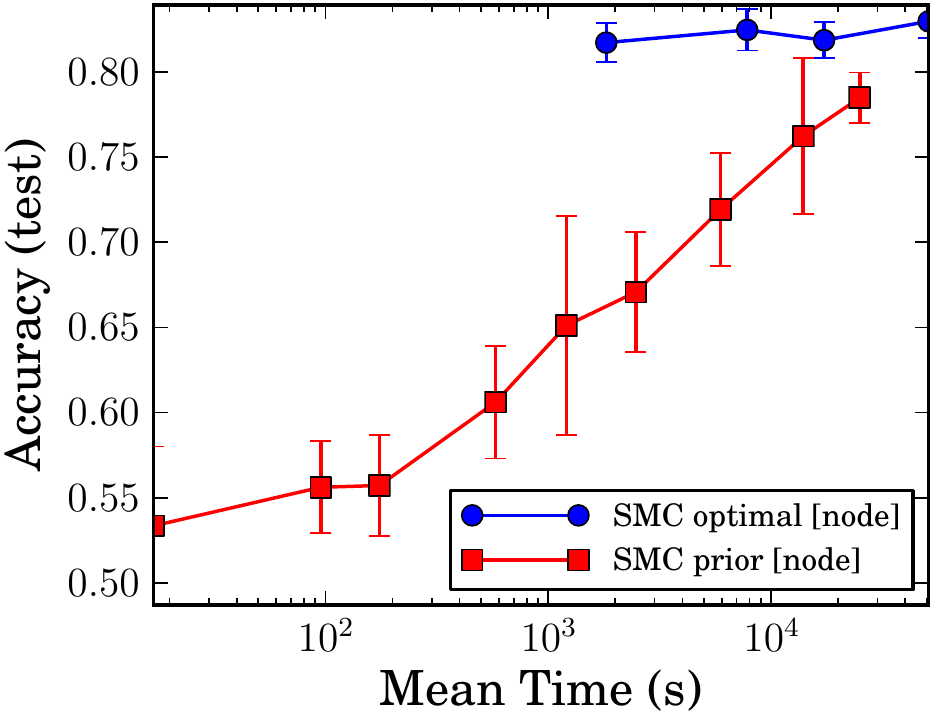}
        \end{subfigure}
        \begin{subfigure} 
                \centering
                \includegraphics[width=\sfigwidth]{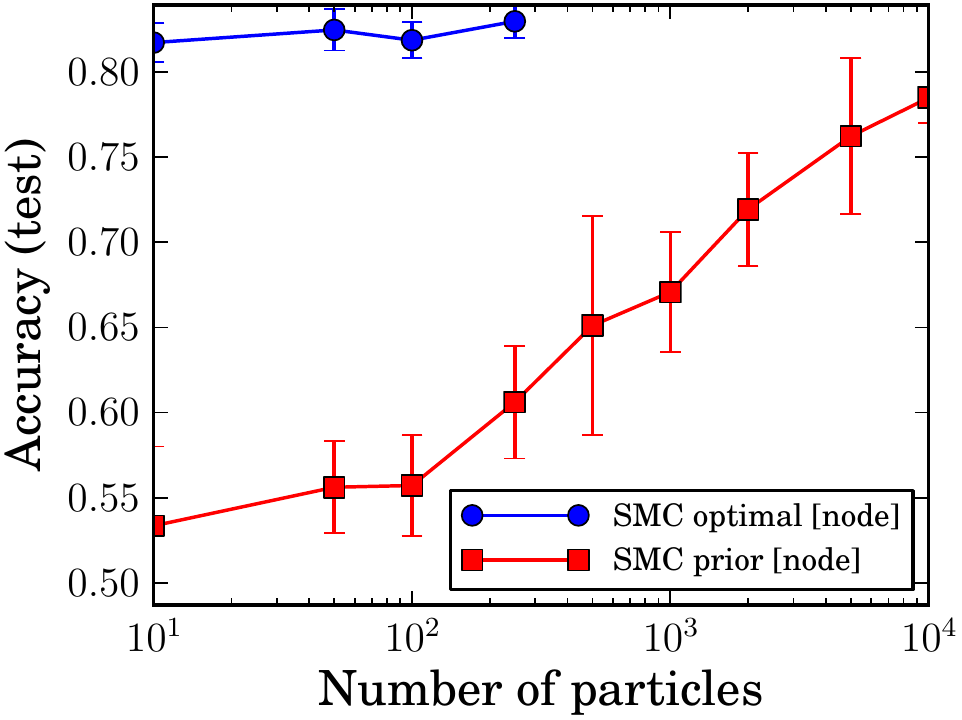}
        \end{subfigure}
              \caption{%
          Results on \madelon\ dataset: %
          The top and bottom rows display $\log p(y|\bx)$ and accuracy on the test data against runtime (left) and the number of particles (right)  respectively. The blue circles and red squares represent \posterior\ and \prior\ proposals respectively.}
        \label{fig:logprobtest:acctest:time:p:madelon}
\end{figure}

The setup is identical to the previous section. The results are shown in Figure \ref{fig:logprobtest:acctest:time:p:madelon}. Here, the \posterior\ proposal 
outperforms the \prior\ proposal in both the columns, requiring fewer particles as well as outperforming the \prior\ proposal for a given computational budget. While this dataset is atypical %
(only $4\%$ of the features are relevant), it illustrates a potential vulnerability of the \prior\ proposal to irrelevant features. 

\subsubsection{Effect of the number of islands}\label{sec:expt:islands}

\begin{figure}
        \centering
        \begin{subfigure}
                \centering
                \includegraphics[width=\sfigwidth]{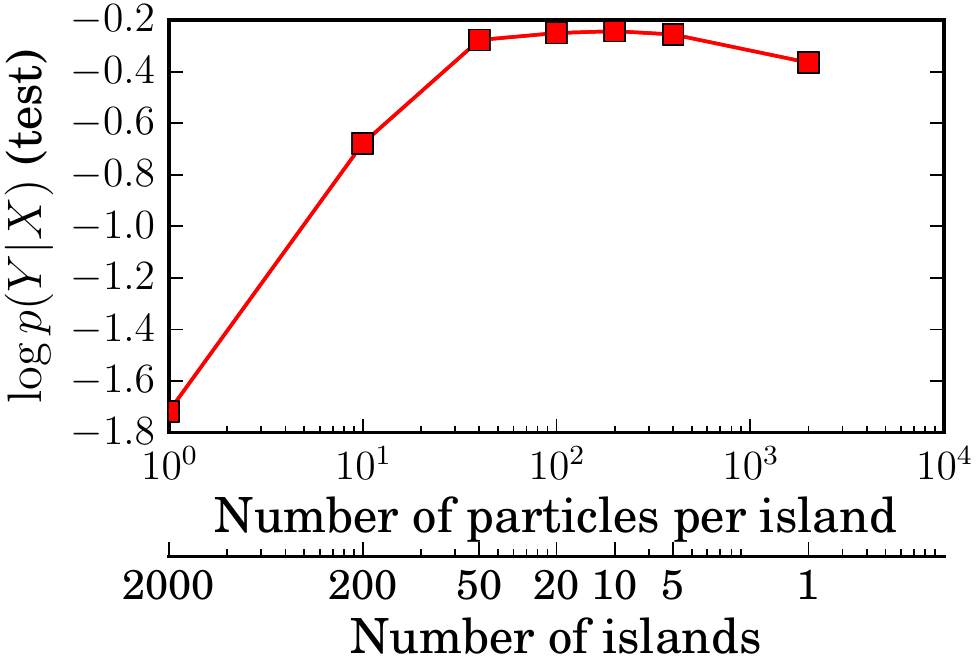}
        \end{subfigure}
                \begin{subfigure}
                \centering
                \includegraphics[width=\sfigwidth]{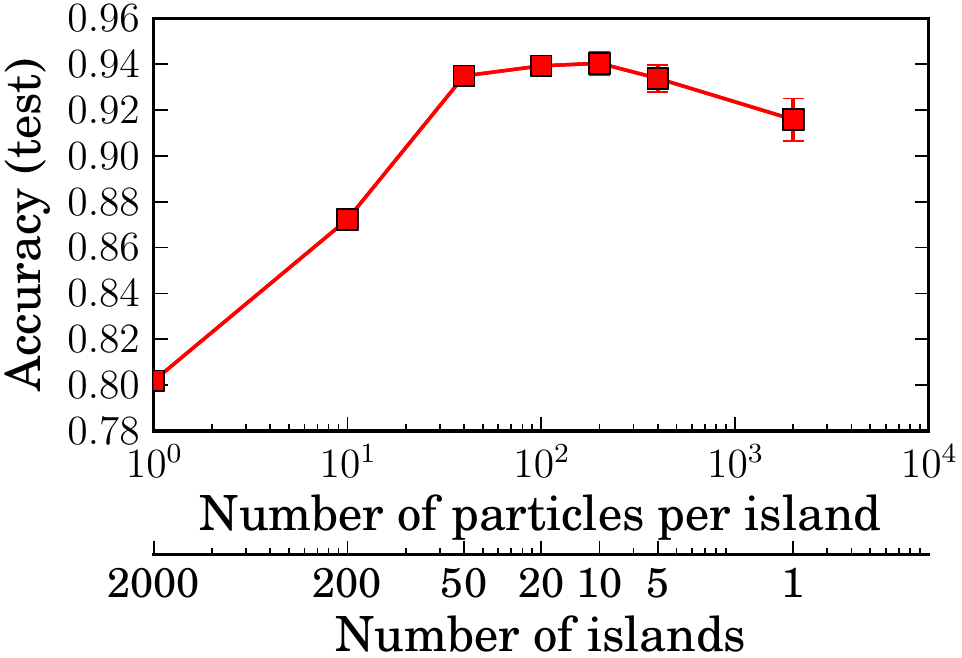}
        \end{subfigure}
        \caption{%
        Results on \pendigits:  Test $\log p(y|\bx)$ (left) and accuracy (right) vs $\nislands$ and $\nparts/\nislands$ for fixed $\nparts=2000$.} %
                \label{fig:logprobtest:acctest:n_islands}
\end{figure}

Averaging the results of several independent particle filters (aka \emph{islands}) is a way to reduce variance at the cost of bias, compared with running a single, larger filter.  In the asymptotic regime, this would not make sense, but as we will see, performance is improved with multiple islands, suggesting we are not yet in the asymptotic regime.
In this experiment, we evaluate the effect of the number of islands on the test performance of the \prior\ proposal. We fix the total number of particles to $2000$ and vary $\nislands$, the number of islands (and hence, the number of particles per island). Note that all the islands operate on the entire dataset unlike \emph{bagging}. %
 Here, we present results only on the \pendigits\ dataset (see Appendix~\ref{sec:islands:magic} for results on the \magic\ dataset). %
 The results are shown in Figure \ref{fig:logprobtest:acctest:n_islands}. We observe that (i) the test performance drops sharply if we use fewer than 100 particles per island and (ii) when $\nparts/\nislands\geq100$, the choices of $\nislands\in[5,100]$ outperform $\nislands=1$. Since the islands are independent,  the computation across islands is `embarrassingly parallelizable'.

\subsection{SMC vs MCMC}\label{sec:expt:smcvsmcmc}

In this experiment, %
we compare the SMC algorithms to the MCMC algorithm proposed by 
\citet{chipman1998bayesian}, which employs four types of Metropolis-Hastings proposals: \emph{grow} (split a leaf node into child nodes), \emph{prune} (prune a pair of leaf nodes belonging to the same parent), \emph{change} (change the decision rule at a node) and \emph{swap} (swap the decision rule of a parent with the decision rule of the child).  In our experiments, we average the MCMC predictions over the trees from all previous iterations.

\begin{figure}
        \centering
        \begin{subfigure} 
                \centering
                \includegraphics[width=\sfigwidth]{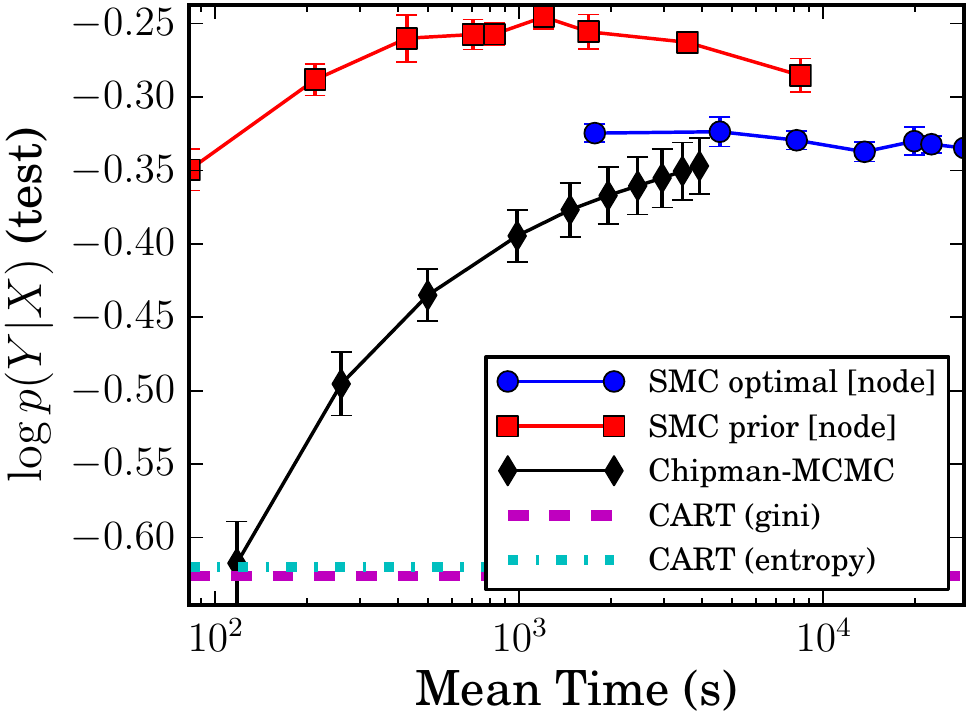}
        \end{subfigure}  
                \begin{subfigure} 
                \centering
                \includegraphics[width=\sfigwidth]{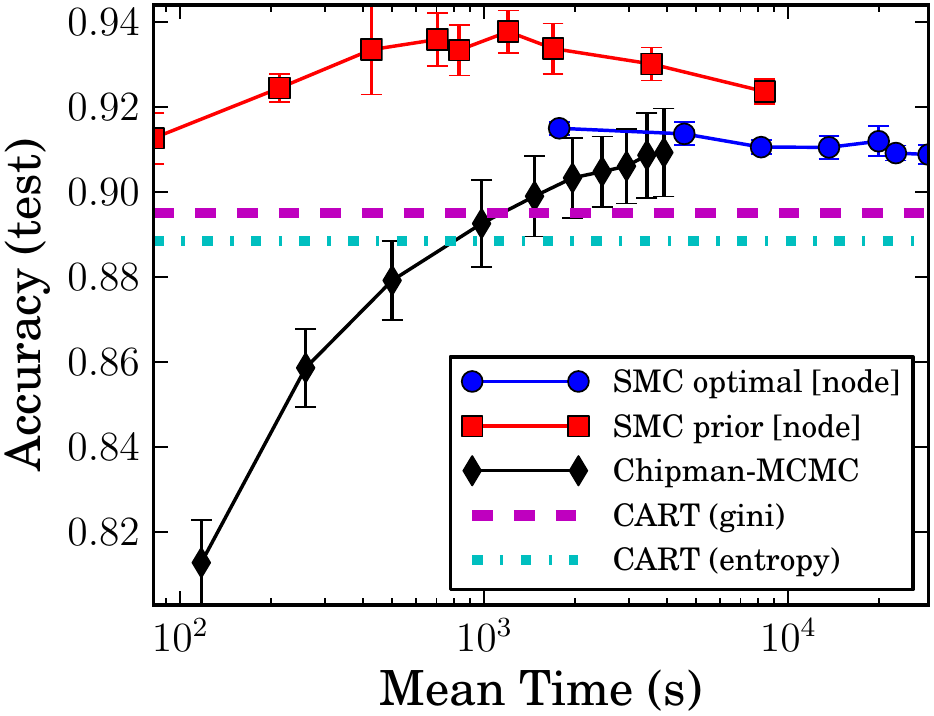}
        \end{subfigure}  
                        \begin{subfigure} 
                \centering
                \includegraphics[width=\sfigwidth]{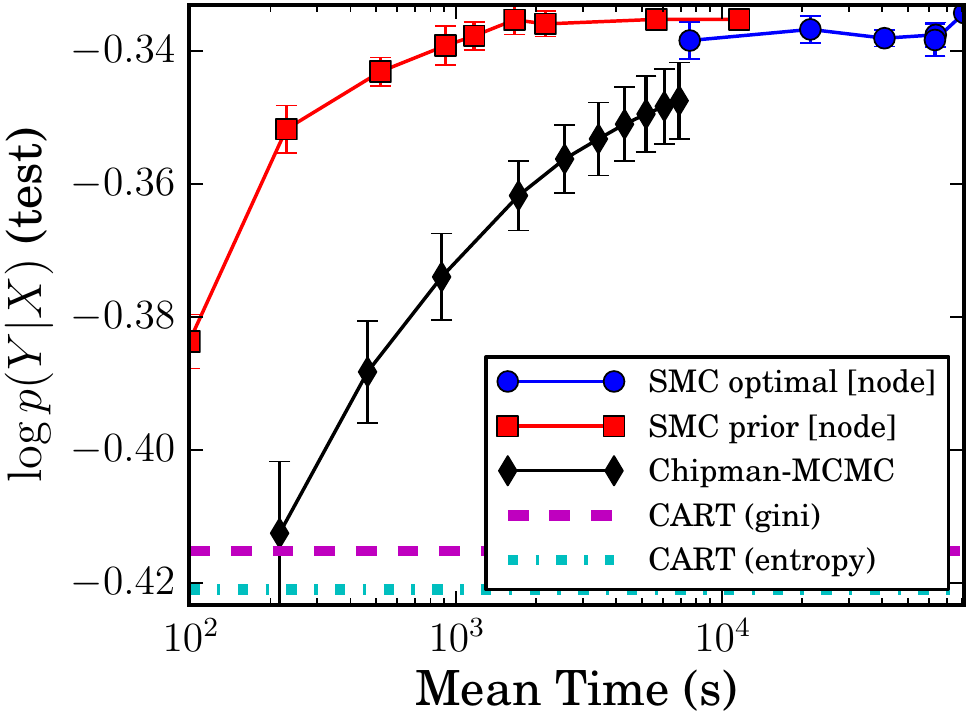}
        \end{subfigure}
                        \begin{subfigure} 
                \centering
                \includegraphics[width=\sfigwidth]{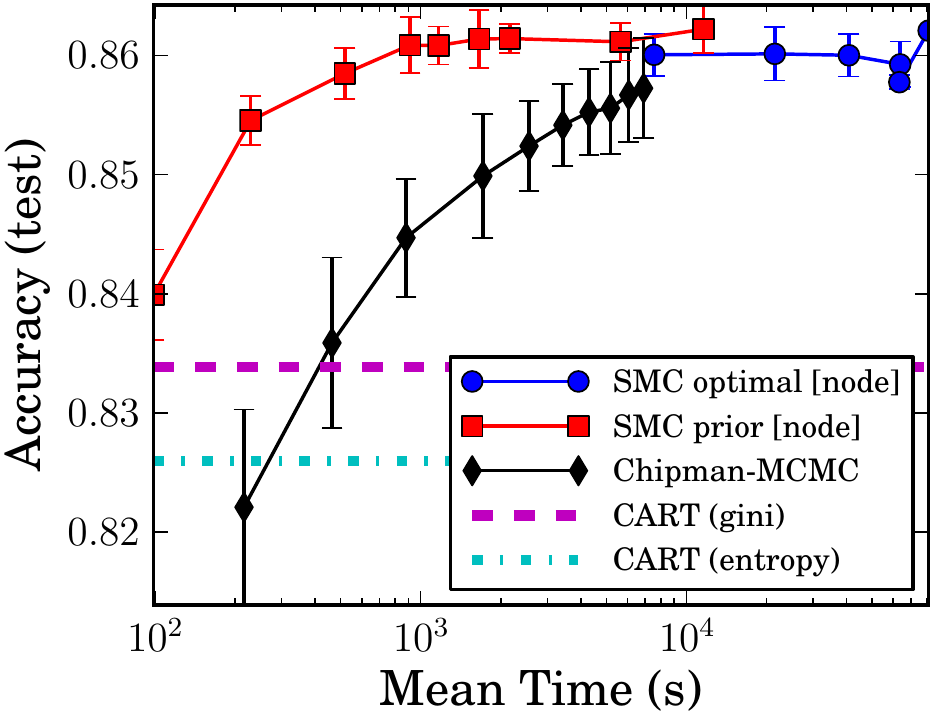}
        \end{subfigure}
        \caption{%
    Results on \pendigits\ (top row), and \magic\ (bottom row). Left column plots test $\log p(y|\bx)$ vs runtime, while right column plots test accuracy vs runtime. The blue cirlces, red squares and black diamonds represent \posterior, \prior\ proposals and MCMC respectively.}
        \label{fig:logprobtest:acctest:time}
\end{figure}

The experimental setup is identical to Section~\ref{sec:smc:design}, except that we fix the number of islands, $\nislands=5$. We vary the number of particles for SMC\footnote{We fix $\nislands=5$ so that the minimum value of $\nparts$ ($=100$) corresponds to $\nparts/\nislands=20$ particles per island. Further improvements could be obtained by `adapting' $\nislands$ to $\nparts$ as discussed in Section \ref{sec:expt:islands}.} and the number of iterations for MCMC %
 and plot the log predictive probability and accuracy on the test data as a function of runtime. 
In Figure~\ref{fig:logprobtest:acctest:time},
we observe that SMC (\prior, \nodewise) is roughly two orders of magnitude faster than MCMC while achieving similar predictive performance on \pendigits\ and \magic\ datasets. Although the exact speedup factor depends on the dataset in general, we have observed  that \textbf{SMC (\prior, \nodewise)  is at least  an order of magnitude faster than MCMC}.   The SMC runtimes in  Figure \ref{fig:logprobtest:acctest:time} are recorded by running the $\nislands$ islands in a serial fashion. As discussed in Section \ref{sec:expt:islands}, one could parallelize the computation leading to an additional speedup by a factor of $\nislands$. In the \pendigits\ dataset, the performance of \prior\ proposal seems to drop as we increase $\nparts$ beyond 2000.  However, the marginal likelihood on the training data increases with $\nparts$ (see Appendix~\ref{sec:expt:marginal}). We believe that the deteriorating performance is due to model misspecification (axis-aligned decision trees are hardly the `right' model for handwritten digits) rather than the inference algorithm itself: `better' Bayesian inference in a misspecified model might lead to a poorer solution (see \citep{minka2000bayesian} for a related discussion).

To evaluate the sensitivity of the trends above to the hyper parameters $\alpha, \alpha_s,\beta_s$, we systematically varied the  values of these hyper parameters and repeated the experiment. The results are qualitatively similar. See Appendix~\ref{sec:expt:sensitivity} for additional information. 

\subsection{SMC vs other existing approaches}%
The goal of these experiments was to verify that our SMC approximation performed as well as the ``gold standard'' MCMC algorithms most commonly used in the Bayesian decision tree learning setting.  Indeed, our results suggest that, for a fraction of the computational budget, we can achieve a comparable level of accuracy.  In this final experiment, we re-affirm that the Bayesian algorithms are competitive in accuracy with the classic CART algorithm.  (There are many other comparisons that one could pursue and other authors have already performed such comparisons.  E.g., \citet{taddy2011} demonstrated that their tree structured models yield similar performance as Gaussian processes and random forests.)
We used the CART implementation provided by \emph{scikit-learn} \citep{scikit-learn} with two criteria: \emph{gini purity} and \emph{information gain} and set $\texttt{min\_samples\_leaf}=10$ (minimum number of data points at a leaf node).\footnote{Lower values ($\texttt{min\_samples\_leaf}=1,5$) tend to yield slightly higher test accuracies (comparable to SMC and MCMC) but much lower predictive probabilities.}  
In addition, we performed Laplacian smoothing on the probability estimates from CART using the same $\alpha$ as for the Bayesian methods. 
Our Python implementation of SMC takes about 50-100x longer to achieve the same test accuracy as the highly-optimized implementation of CART.  For this reason, we plot CART accuracy as a horizontal bar.
The accuracy and log predictive probability on test data are shown in Figure \ref{fig:logprobtest:acctest:time}. 
The Bayesian decision tree frameworks achieve similar (or better) test accuracy to CART, and outperform CART significantly in terms of the predictive likelihood.  SMC delivers the benefits of having an approximation to the posterior, but in a fraction of the time required by existing MCMC methods.

\section{Discussion and Future work}
\label{sec:discussion}

We have proposed a novel class of Bayesian inference algorithms for decision trees, based on the sequential Monte Carlo framework.  The %
algorithms mimic classic top-down algorithms for learning decision trees, but use ``local'' likelihoods along with resampling steps to guide tree growth.  We have shown good computational and statistical performances, especially compared with a state-of-the-art MCMC inference algorithm.  Our algorithms are easier to implement than their MCMC counterparts, whose efficient implementations require sophisticated book-keeping.

We have also explored various design choices leading to different SMC algorithms.  We have found that expanding too many nodes simultaneously degraded performance, and more sophisticated ways of choosing nodes surprisingly did not improve performance.  %
Finally, while the one-step \optimal\ proposal often required fewer particles to achieve a given accuracy, it was significantly more computationally intensive than the \prior\ proposal, leading to a less efficient algorithm overall on datasets with few irrelevant input dimensions.  As the number of irrelevant dimensions increased the balance tipped in favour of the \optimal\ proposal.  An interesting direction of exploration is to devise some way to interpolate between the \prior\ and \optimal\ proposals, getting the best of both worlds.

The model underlying this work assumes that the data is explained by a single tree.
In contrast, many uses of decision trees, e.g., random forests, bagging, etc., can be interpreted as working within a model class where the data is explained by a collection of trees. 
Bayesian additive regression trees (BART) \citep{Chipman10} are such a model class.
Prior work has considered MCMC techniques for posterior inference \citep{Chipman10}.  A significant but important extension of this work would be to tackle %
 additive combinations of trees, potentially in a way that continues to mimic classic algorithms.

Finally, in order to more closely match existing work in Bayesian decision trees, we have used a prior over decision trees that depends on the input data $\bX$.  This has the undesirable side-effect of breaking exchangeability in the model, making it incoherent with respect to changing dataset sizes and to working with online data streams.  One solution is to use an alternative prior for decision trees, e.g., based on the Mondrian process \citep{RoyTeh2009a}, whose projectivity would re-establish exchangeability while allowing for efficient posterior \emph{computations} that depend on data.%

\section*{Acknowledgments}

We would like to thank 
Charles Blundell, 
Arnaud Doucet, 
David Duvenaud, 
Jan Gasthaus, 
Hong Ge,
Zoubin Ghahramani,
and
James Robert Lloyd
for helpful discussions and feedback on drafts.
DMR is supported by a Newton International Fellowship and Emmanuel College.
BL and YWT gratefully acknowledge generous funding from the Gatsby Charitable Foundation.

\bibliography{tree-smc}
\bibliographystyle{abbrvnat}

\newpage
\appendix
\input{arxiv-supplementarytext.tex}

\end{document}

%% file: macros.tex
\newcommand{\real}{\mathrm{I\kern-0.175em R}}

\def\nn{\nonumber}

\def\O{\mathcal{O}}

\def\T{\mathcal{T}}

\def\bx{x}

\def\bz{\bm{z}}

\def\bX{X}

\def\indicator{\mathds{1}}

\def\pp{p}

%% file: arxiv-supplementarytext.tex
\appendix

\section{SMC algorithm}\label{sec:alg:smc}
\begin{algorithm}[h!] 
  \caption{SMC for Bayesian decision tree learning}   
\label{alg:smc}                       
\begin{algorithmic}                 
    \STATE Inputs: Training data $(\bX, Y)$ \\
                             \hspace{12.5mm}Number of particles $\nparts$ 
    \STATE Initialize: 
    $\T^\cc_0 = E^\cc_0 = \{\epsilon\}$ \\
    \hspace{15.5mm}$\tau^\cc_0=\kappa^\cc_0=\emptyset$ \\
    \hspace{15.5mm}$\wt^\cc_0 = f(Y|\TS^\cc_0)$ \\
    \hspace{15.5mm}$W_0 = \sum_{\Cc} \wt^\cc_0$ \\
    \ \\
    \FOR{$i = 1: \textrm{MAX-STAGES}$}
    \FOR{$\Cc = 1:\nparts$}
    \STATE Sample $\TS^\cc_i$ from $\Kr_i(\cdot \given \TS^\cc_{i-1})$
     \\$\qquad$ where $\TS^\cc_i \defas (\T^\cc_i,\kappa^\cc_i,\tau^\cc_i,E^\cc_i)$
    \STATE Update weights:$\vphantom{\TS^\cc}$ \tiny{(Here $\Pr,\Kr_i$ denote their densities.)}
 \small{\[
\wt^\cc_i 
&=         \frac{ \Pr(\TS^\cc_i ) \, g(Y \given \TS^\cc_i, \bX) }
                     { \Kr_i( \TS^\cc_{i} \given \TS^\cc_{i-1})\, \Pr(\TS^\cc_{i-1} ) } \label{eq:wu1}\\
&= \wt^\cc_{i-1} \frac{\Pr(\TS^\cc_i \given \TS^\cc_{i-1}) }
                     { \Kr_i( \TS^\cc_{i} \given \TS^\cc_{i-1}) }
                      \frac{g(Y \given \TS^\cc_i,\bX) }
                      { g(Y \given \TS^\cc_{i-1},\bX) }
                      \label{eq:wu2}
\]}
    \ENDFOR
    \STATE Compute normalization: 
    $W_i = \sum_{\Cc} \wt^\cc_i$
    \STATE Normalize weights: $(\forall \Cc)\, \wtbar^\cc_i = \wt^\cc_i / W_i$
    \IF{$\bigl(\sum_\Cc (\wtbar^\cc_i)^2\bigr)^{-1} < \textrm{ESS-THRESHOLD}$} 
    \STATE 
    $(\forall \Cc)$ Resample indices $j_\Cc$ from $\sum_{\Ccd} \wtbar^\ccd_i \delta_{\Ccd}$
    \STATE $(\forall \Cc)$ $\TS^\cc_i \gets \TS^{(j_\Cc)}_i$; $\wt^\cc_i\gets W_i/\nparts$
    \ENDIF
    \IF{$(\forall \Cc)\, E^\cc_i = \emptyset$}
    \STATE exit for loop
    \ENDIF
    \ENDFOR
    \RETURN 
    Estimated marginal probability $W_i/\nparts$ and \\
    \hspace{7mm} weighted samples $\{\wt^\cc_i, \T^\cc_i,\kappa^\cc_i, \tau^\cc_i\}_{\Cc=1}^{\nparts}$.
\end{algorithmic}
\end{algorithm}

\section{Effect of SMC proposal and expansion strategy on test accuracy}\label{sec:expt:smc:acctest}
\begin{figure}[h]
        \centering
        \begin{subfigure} 
                \centering
                \includegraphics[width=\figwidth]{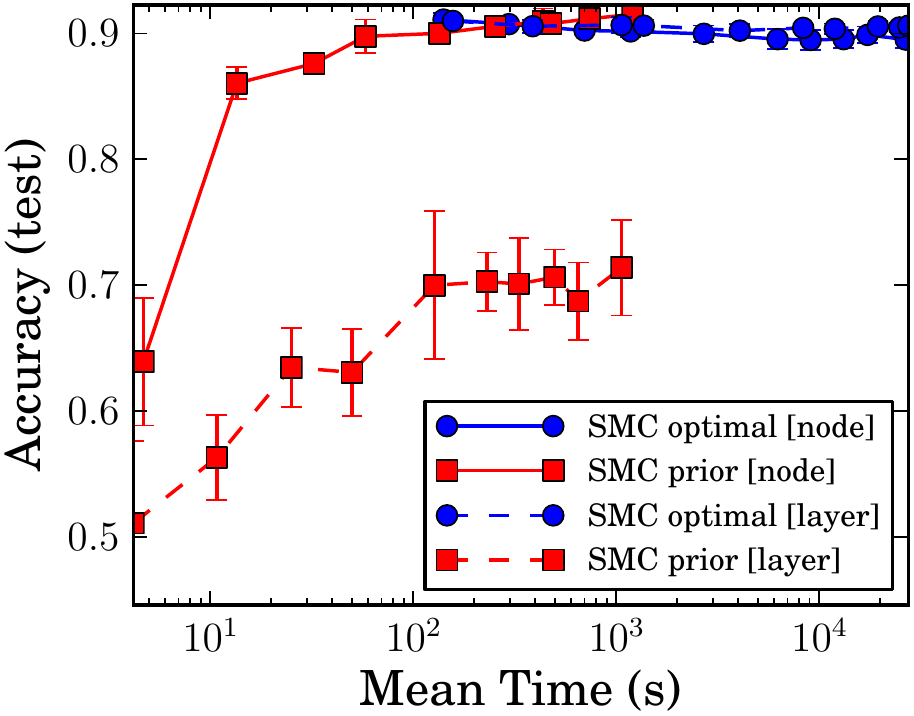}
        \end{subfigure}  \begin{subfigure} 
                \centering 
                \includegraphics[width=\figwidth]{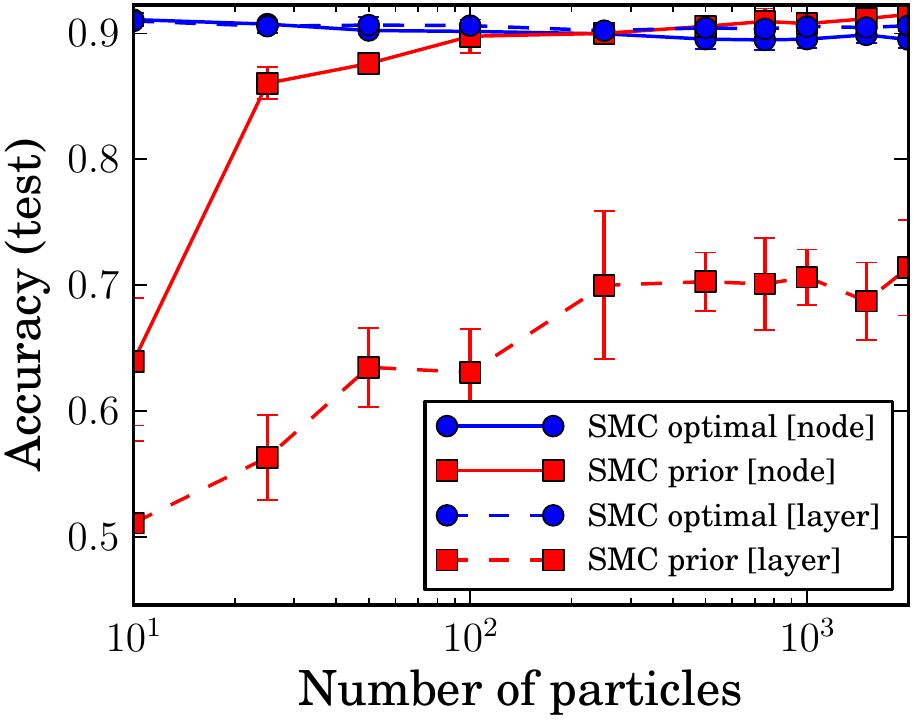}
        \end{subfigure}
        ~
                \begin{subfigure} 
                \centering
                \includegraphics[width=\figwidth]{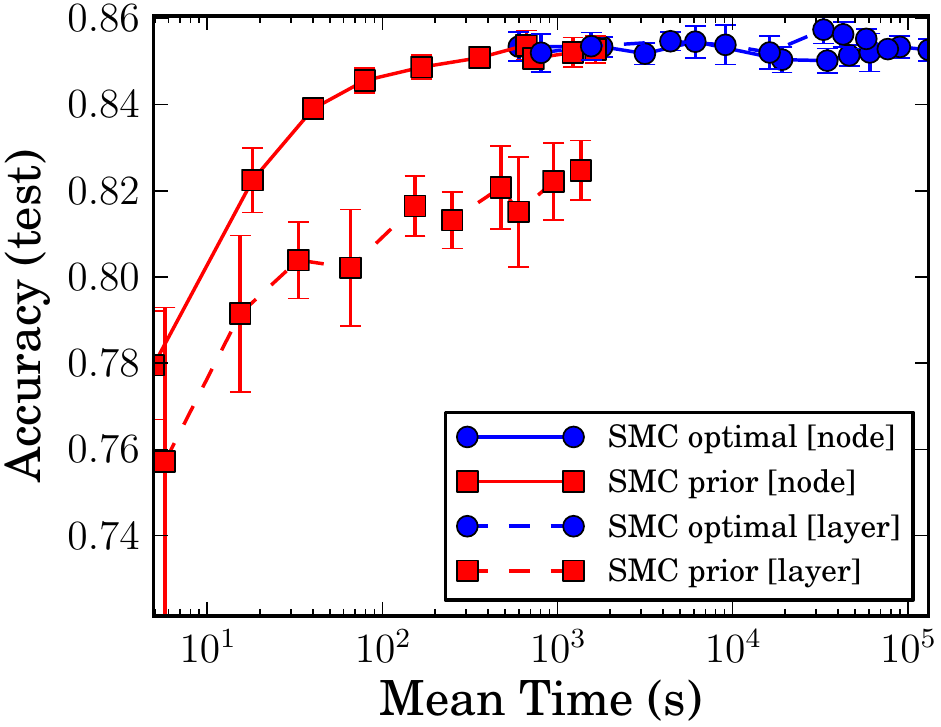}
        \end{subfigure}
        \begin{subfigure} 
                \centering
                \includegraphics[width=\figwidth]{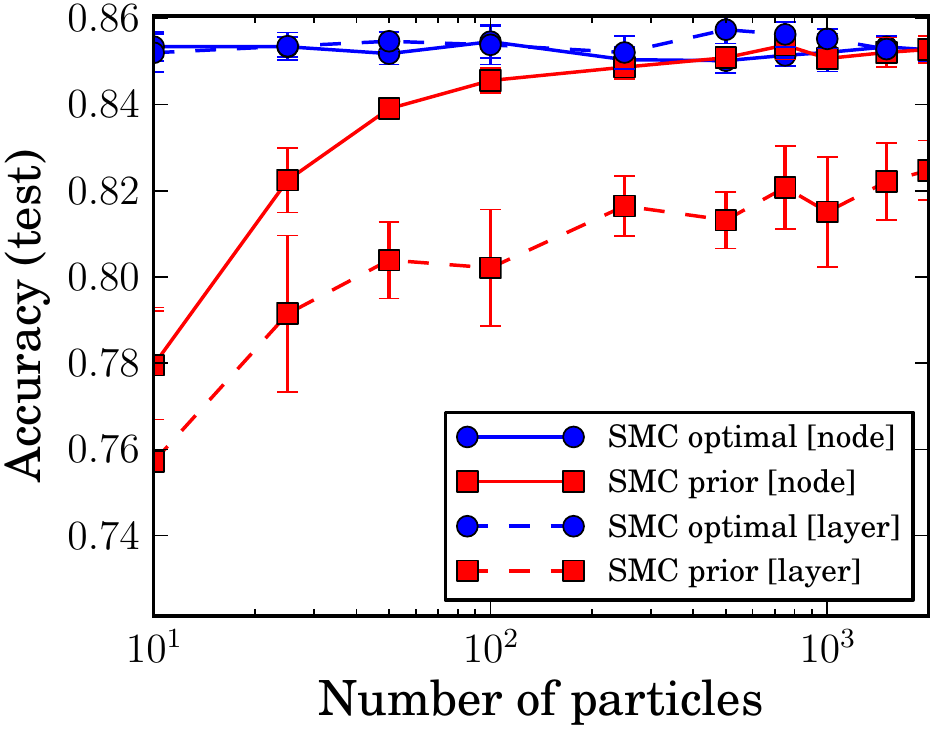}
        \end{subfigure}
        \caption{%
    Results on \pendigits\ (top), and \magic\ (bottom). Left column plots test accuracy vs runtime, while right column plots test accuracy vs number of particles. The blue circles and red squares represent \posterior\ and \prior\ proposals respectively. The solid and dashed lines represent \nodewise\ and \layerwise\ proposals respectively. }
        \label{fig:acctest:time:p}
\end{figure}

The results are shown in Figure \ref{fig:acctest:time:p}.

\section{Effect of the number of islands: \magic\ dataset}\label{sec:islands:magic}
\begin{figure}[h]
        \centering
        \begin{subfigure}
                \centering
                \includegraphics[width=\sfigwidth]{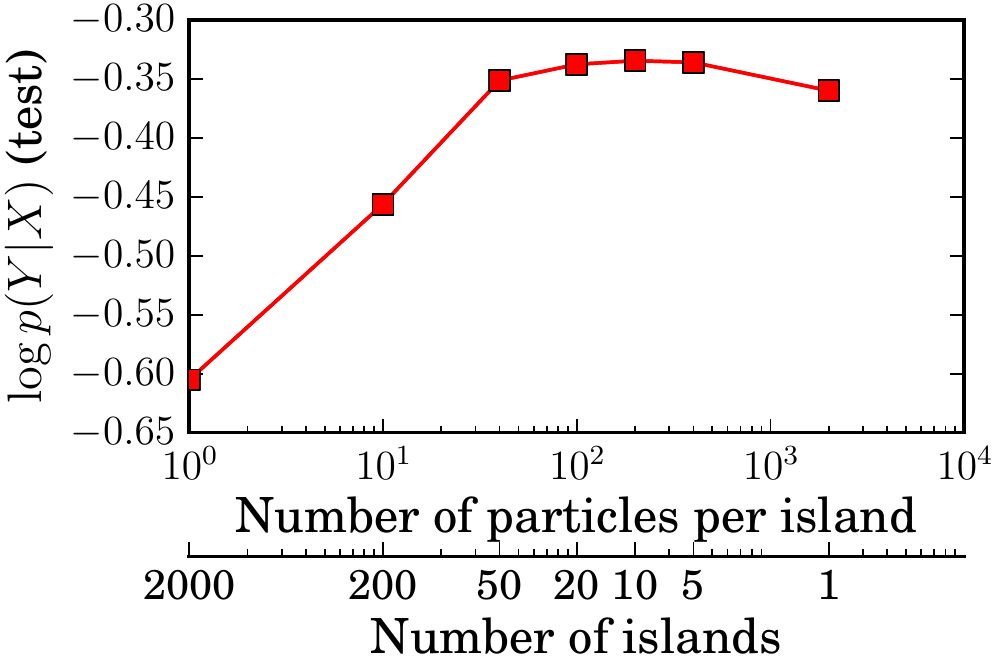}
        \end{subfigure}
                \begin{subfigure}
                \centering
                \includegraphics[width=\sfigwidth]{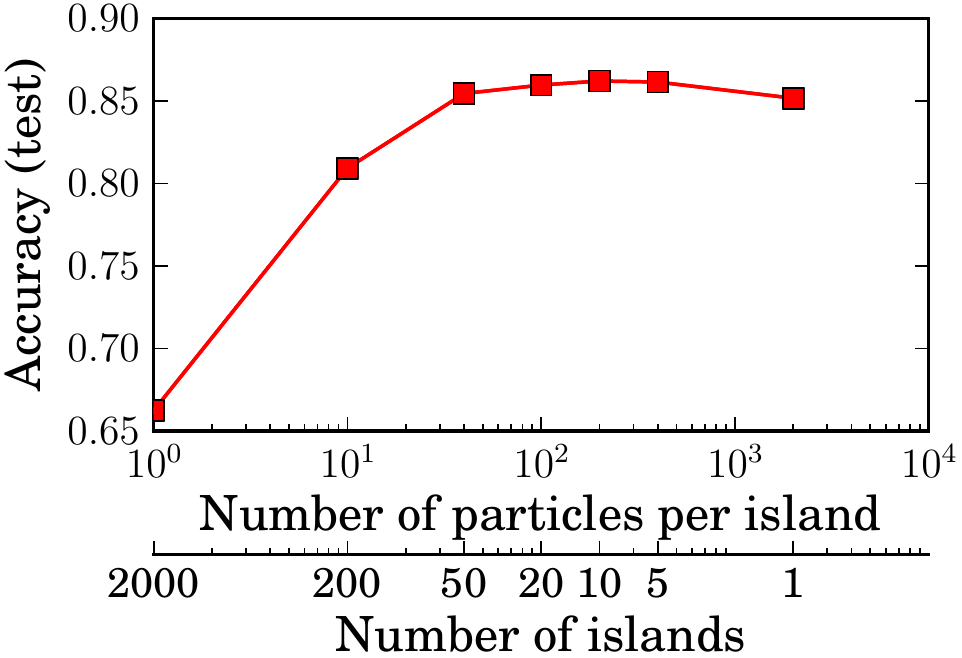}
        \end{subfigure}
        \caption{%
        Results on \magic:  Test $\log p(y|\bx)$ (left) and accuracy (right) vs $\nislands$ and $\nparts/\nislands$ for fixed $\nparts=2000$.}                 \label{fig:logprobtest:acctest:n_islands:magic}
\end{figure}

The results are shown in Figure \ref{fig:logprobtest:acctest:n_islands:magic}. 

\section{Marginal likelihood}\label{sec:expt:marginal}
\begin{figure}[h]
        \centering
        \begin{subfigure} 
                \centering
   \includegraphics[width=\figwidth]{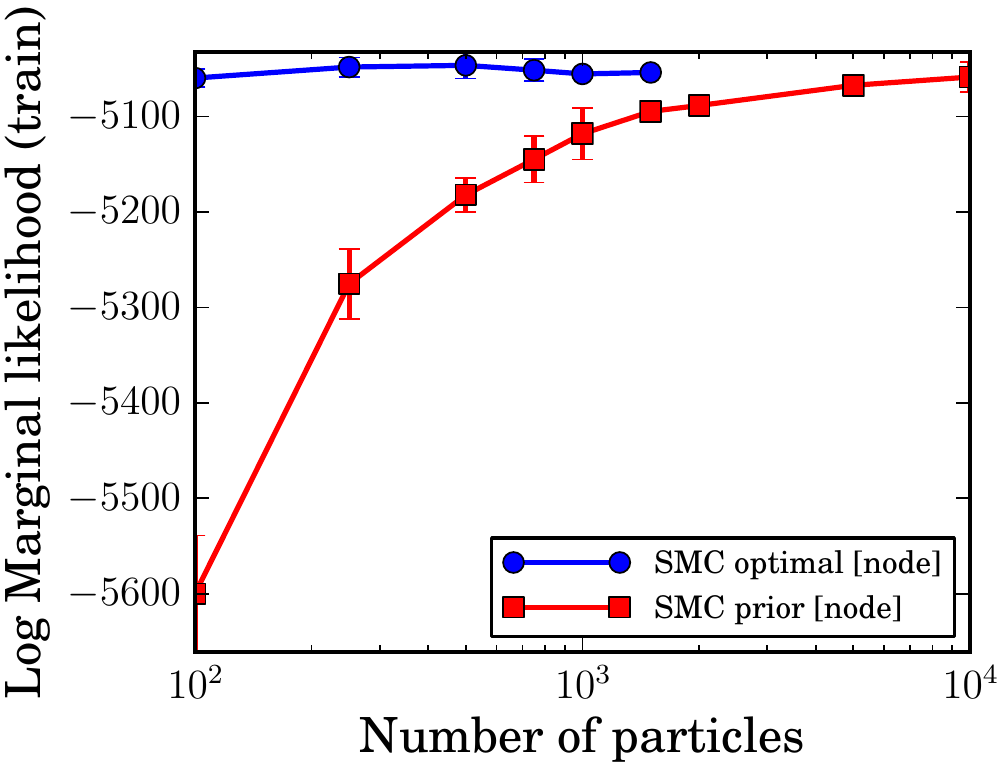}
        \end{subfigure}  \begin{subfigure} 
                \centering
                \includegraphics[width=\figwidth]{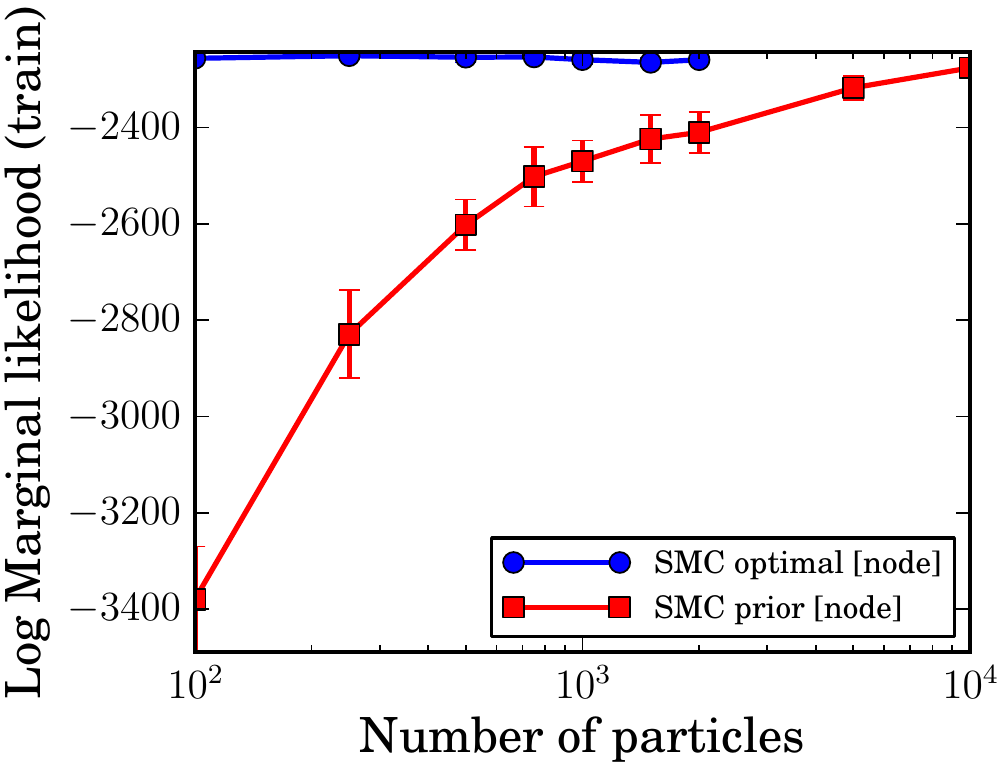}
        \end{subfigure}
              \caption{%
              Results on \pendigits\ (left), and \magic\ (right). Mean log marginal likelihood (i.e., mean $\log p(Y|\bX)$ for training data averaged across 10 runs) vs number of particles. The blue circles and red squares represent \posterior\ and \prior\ proposals respectively.}
        \label{fig:pvslogprobtrain:2}
\end{figure}

The log marginal likelihood of the training data for different proposals is shown in Figure \ref{fig:pvslogprobtrain:2}. As the number of particles increases, the log marginal likelihood of \prior\ and \posterior\ proposals converge to the same value (as expected). 

\section{Sensitivity of results to choice of hyperparameters}\label{sec:expt:sensitivity}
\begin{figure}[h!]
        \centering
        \begin{subfigure} 
                \centering
                \includegraphics[width=\sfigwidth]{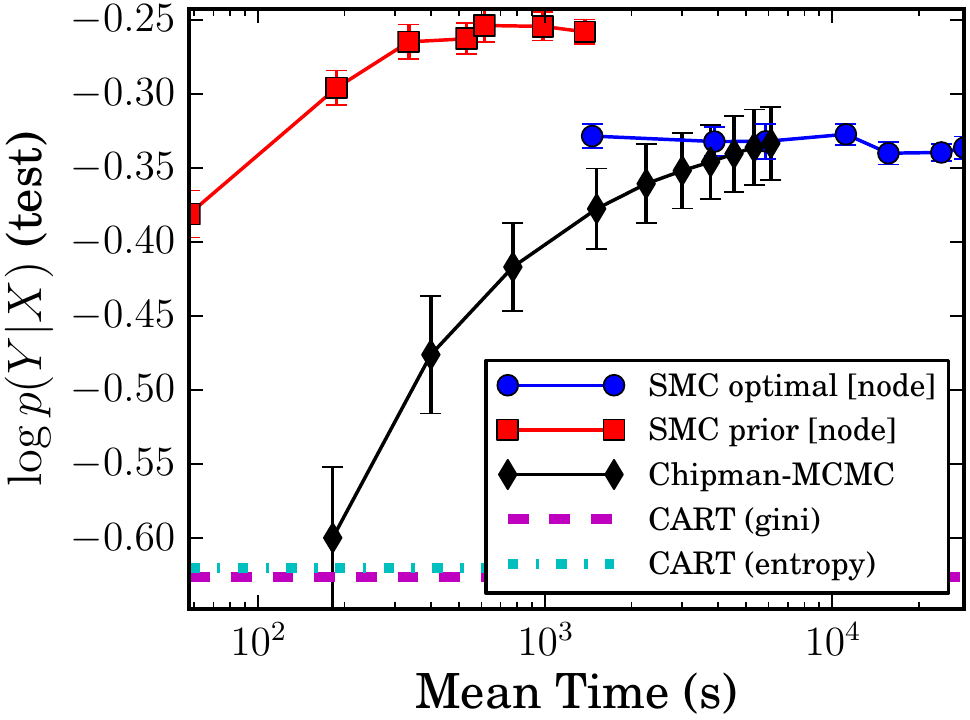}
        \end{subfigure}  
                \begin{subfigure} 
                \centering
                \includegraphics[width=\sfigwidth]{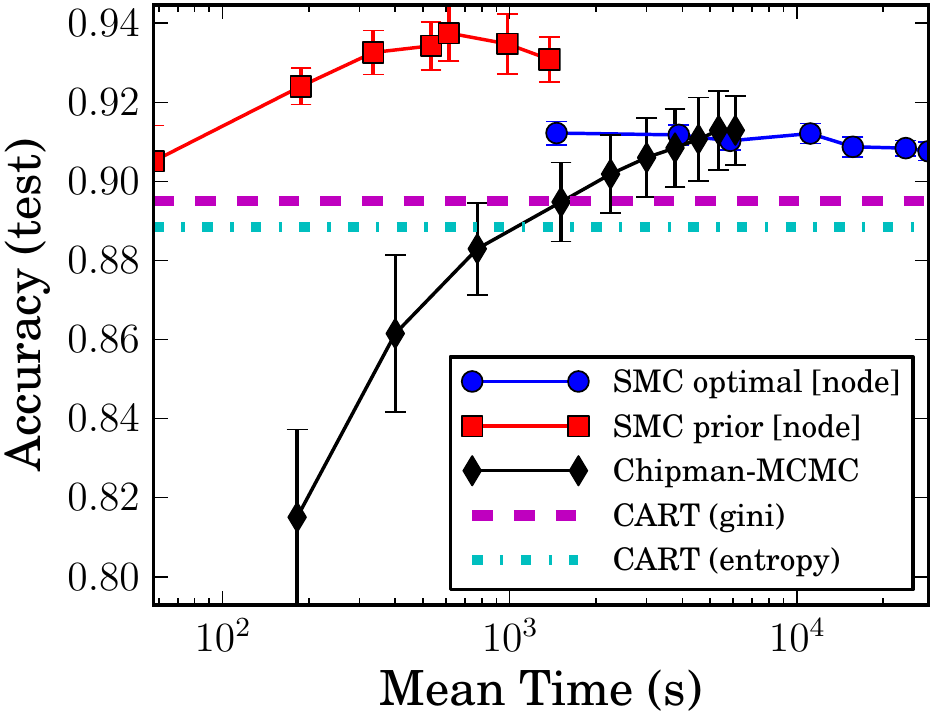}
        \end{subfigure}  
                        \begin{subfigure} 
                \centering
                \includegraphics[width=\sfigwidth]{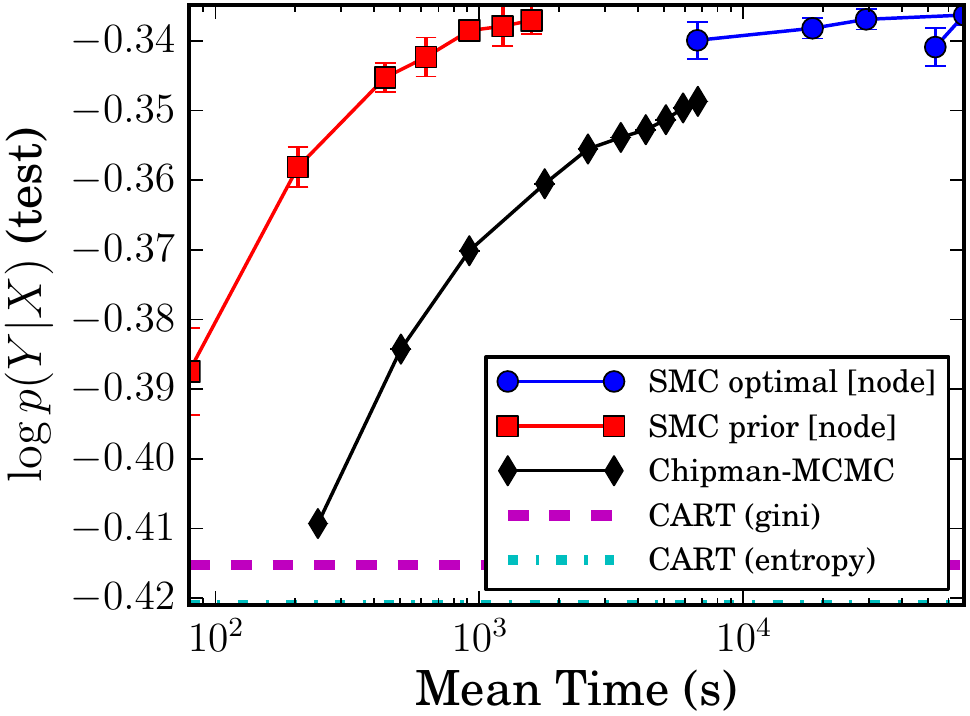}
        \end{subfigure}
                        \begin{subfigure} 
                \centering
                \includegraphics[width=\sfigwidth]{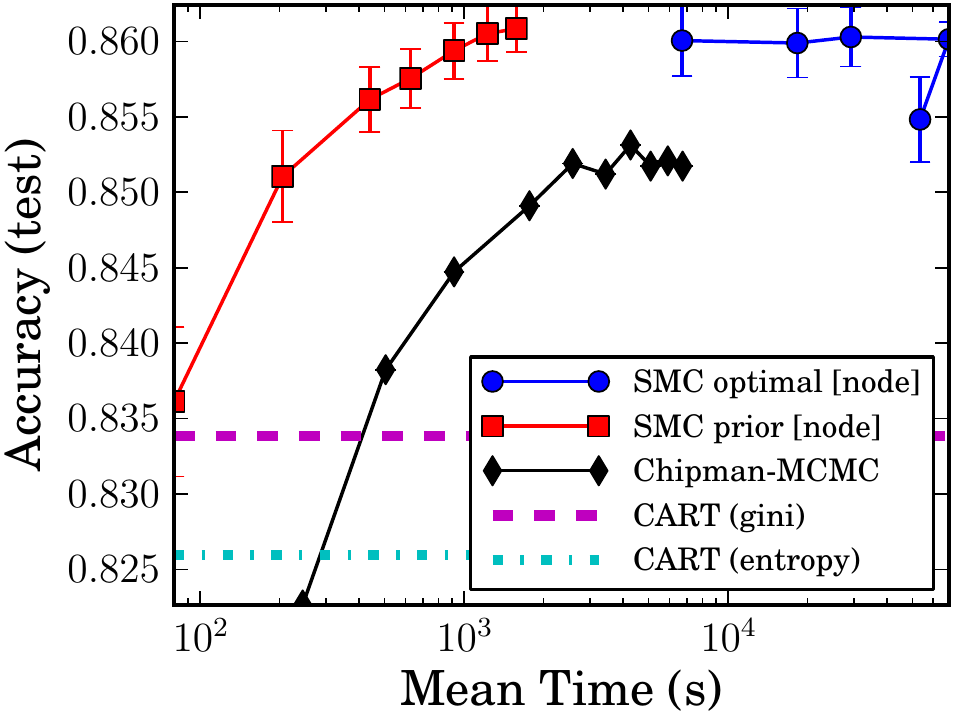}
        \end{subfigure}
           \caption{
          Hyperparameters: $\alpha=5.0,\bm{\alpha_s=0.8},\beta_s=0.5$} (see main text for additional information).
        \label{fig:logprobtest:acctest:time:2}
\end{figure}

\begin{figure}[h!]
        \centering
        \begin{subfigure} 
                \centering
                \includegraphics[width=\sfigwidth]{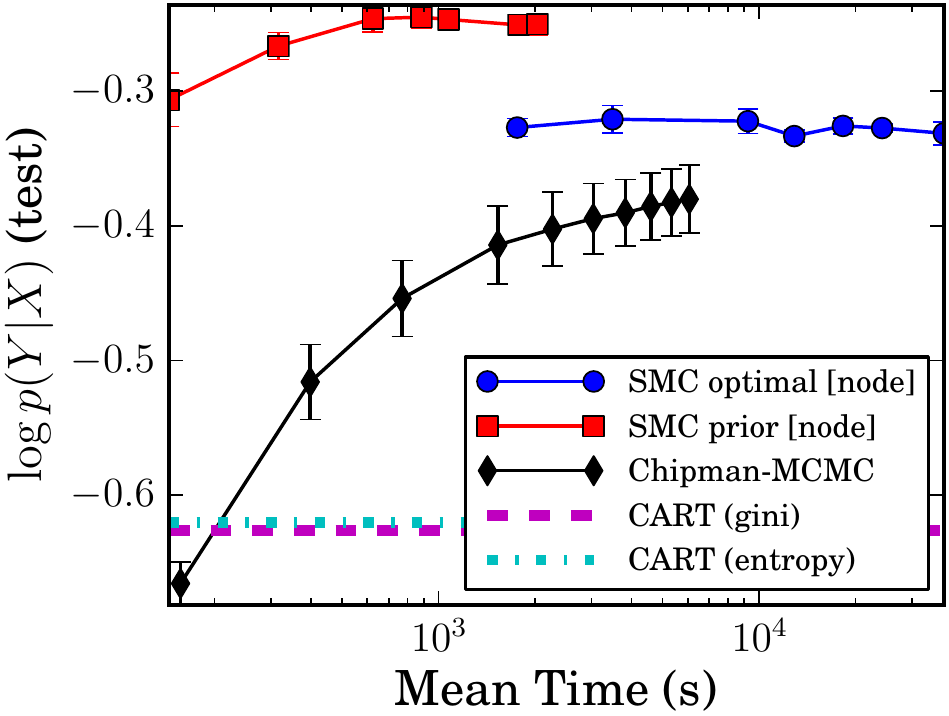}
        \end{subfigure}  
                \begin{subfigure} 
                \centering
                \includegraphics[width=\sfigwidth]{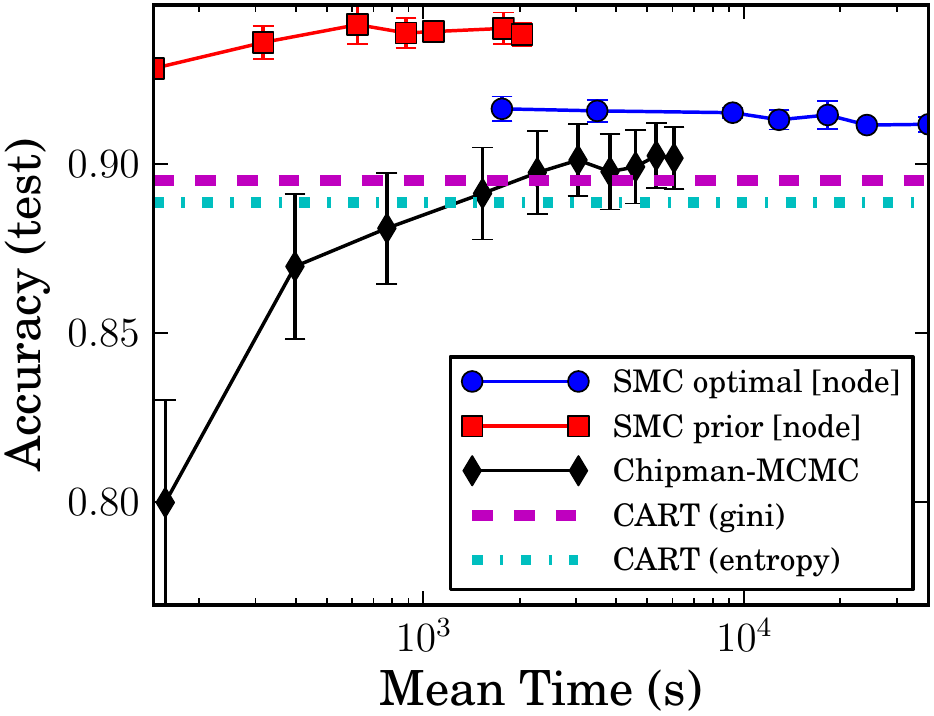}
        \end{subfigure}  
                        \begin{subfigure} 
                \centering
                \includegraphics[width=\sfigwidth]{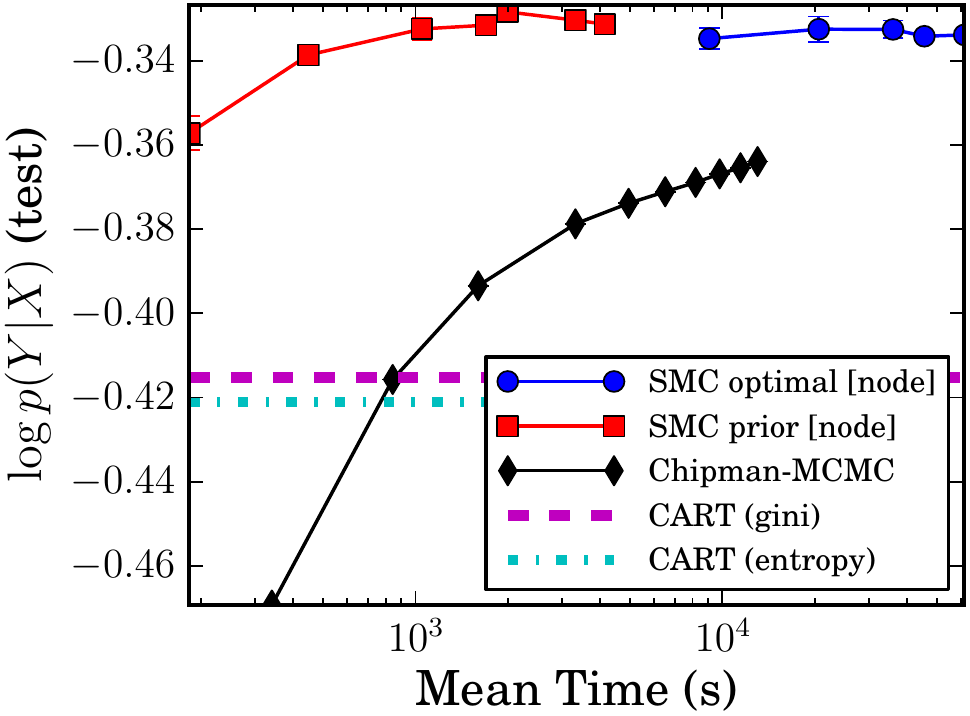}
       \end{subfigure}
                      \begin{subfigure} 
             \centering
            \includegraphics[width=\sfigwidth]{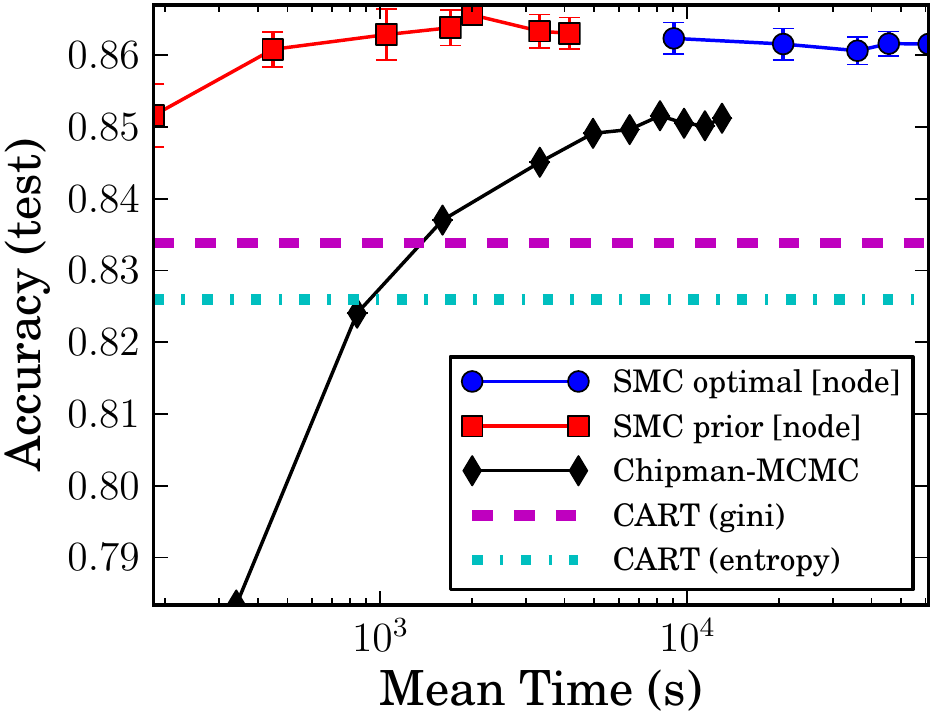}
   \end{subfigure}
   \caption{Hyperparameters: $\alpha=5.0,\alpha_s=0.95,\bm{\beta_s=0.2}$ (see main text for additional information).}
        \label{fig:logprobtest:acctest:time:3}
\end{figure}

\begin{figure}[h!]
        \centering
        \begin{subfigure} 
                \centering
                \includegraphics[width=\sfigwidth]{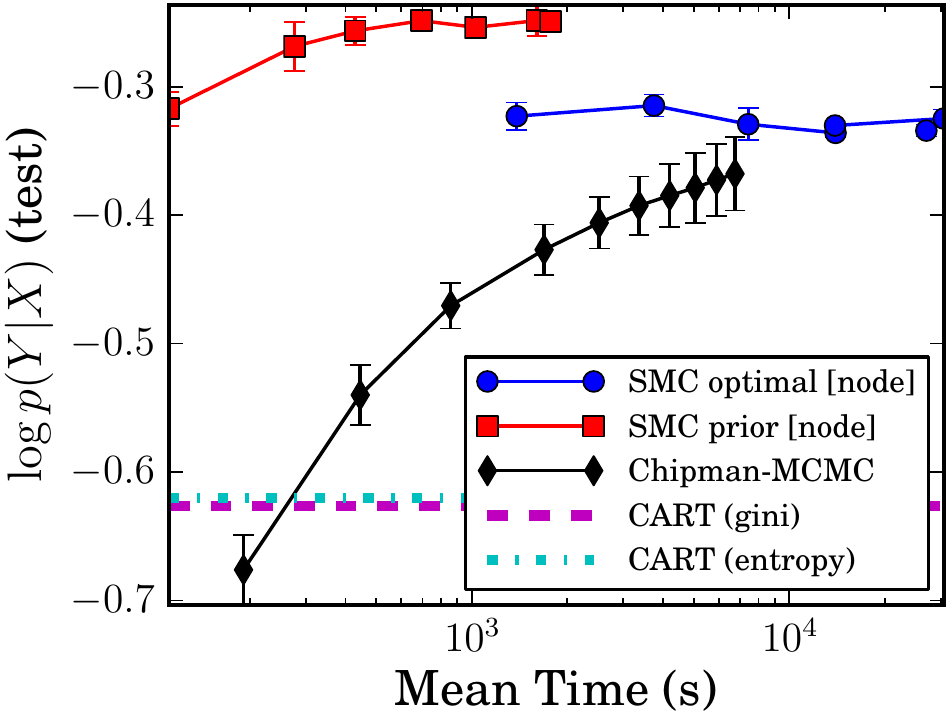}
        \end{subfigure}  
\begin{subfigure} 
                \centering
                \includegraphics[width=\sfigwidth]{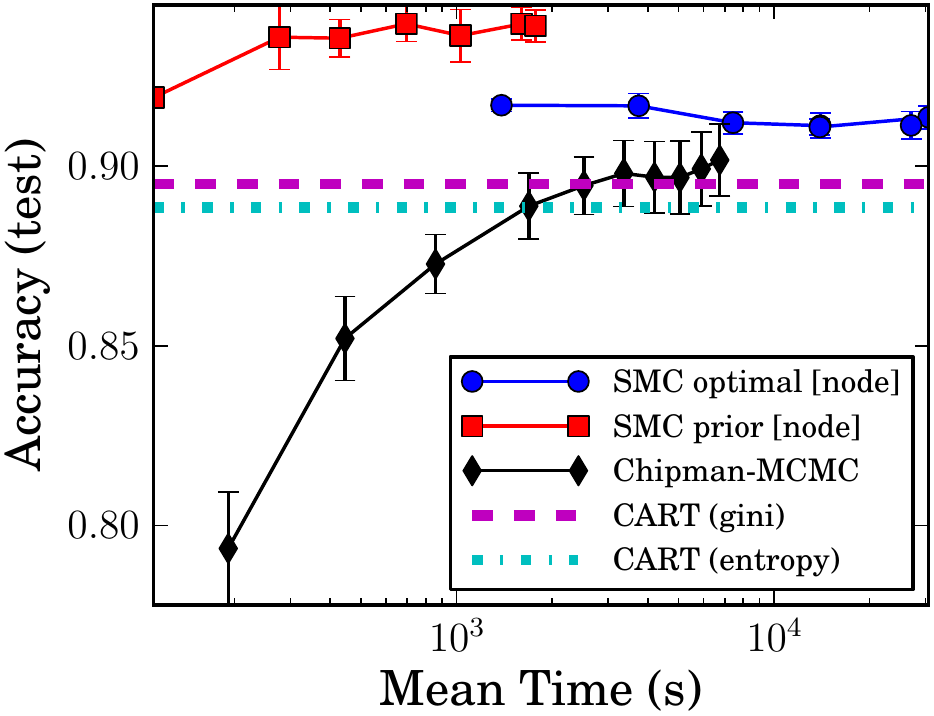}
        \end{subfigure}  
                        \begin{subfigure} 
                \centering
                \includegraphics[width=\sfigwidth]{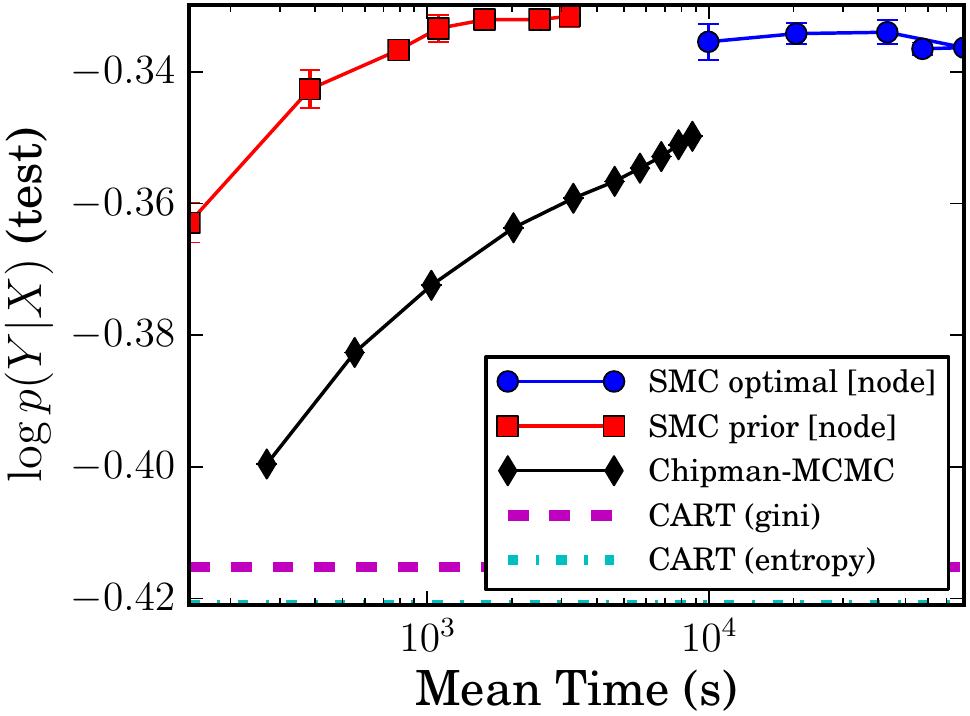}
        \end{subfigure}
                        \begin{subfigure} 
                \centering
                \includegraphics[width=\sfigwidth]{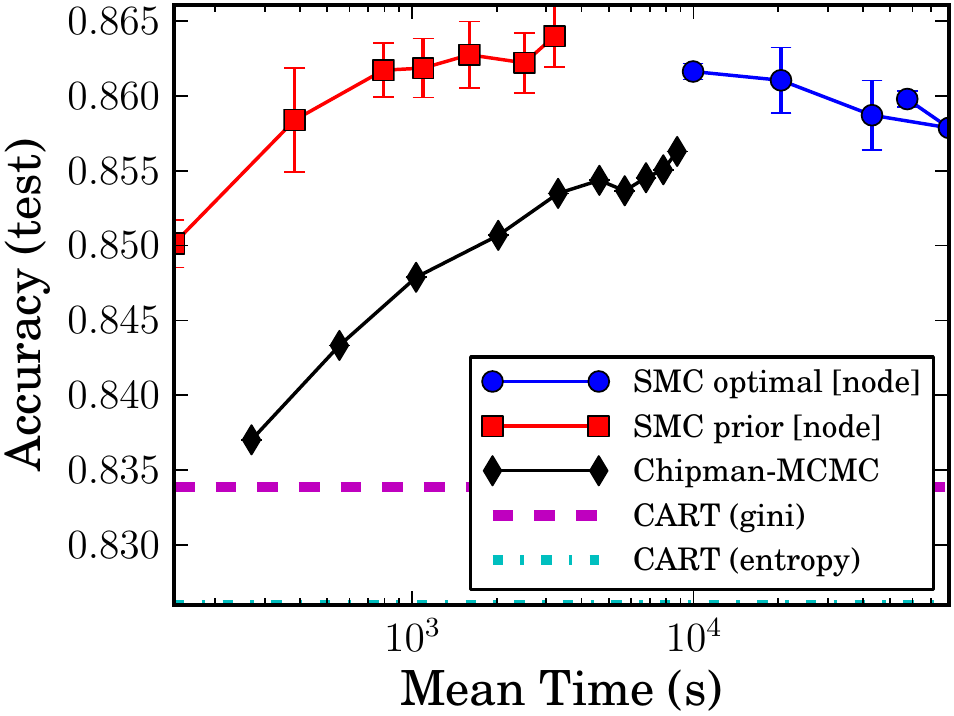}
        \end{subfigure}
           \caption{
        Hyperparameters: $\alpha=5.0,\bm{\alpha_s=0.8,\beta_s=0.2}$ (see main text for additional information). 
           }
        \label{fig:logprobtest:acctest:time:4}
\end{figure}

\begin{figure}[h!]
        \centering
        \begin{subfigure} 
                \centering
                \includegraphics[width=\sfigwidth]{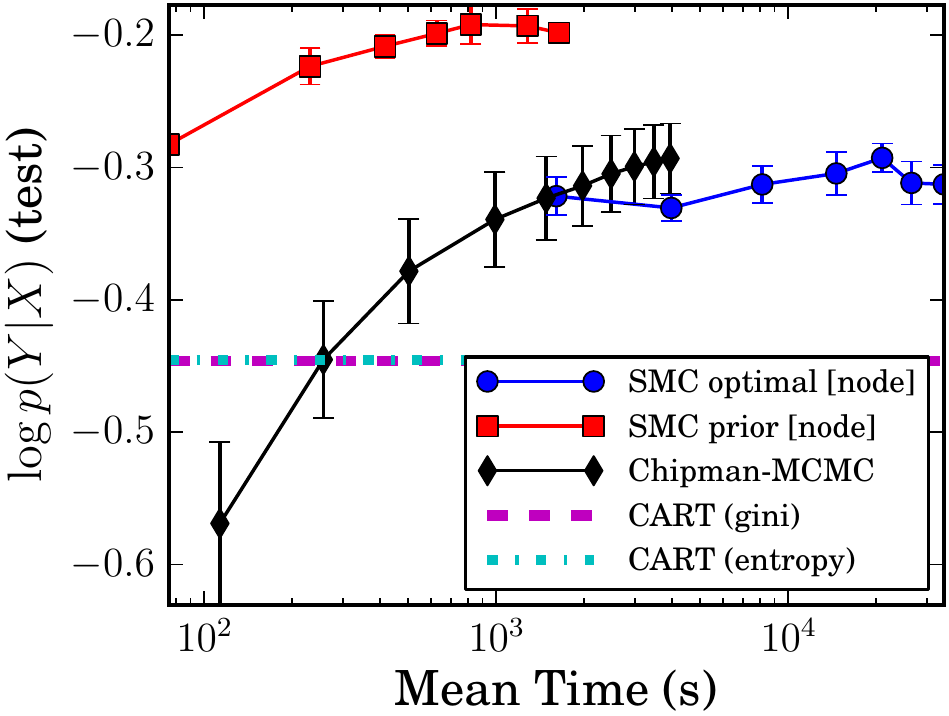}
        \end{subfigure}  
                \begin{subfigure} 
                \centering
                \includegraphics[width=\sfigwidth]{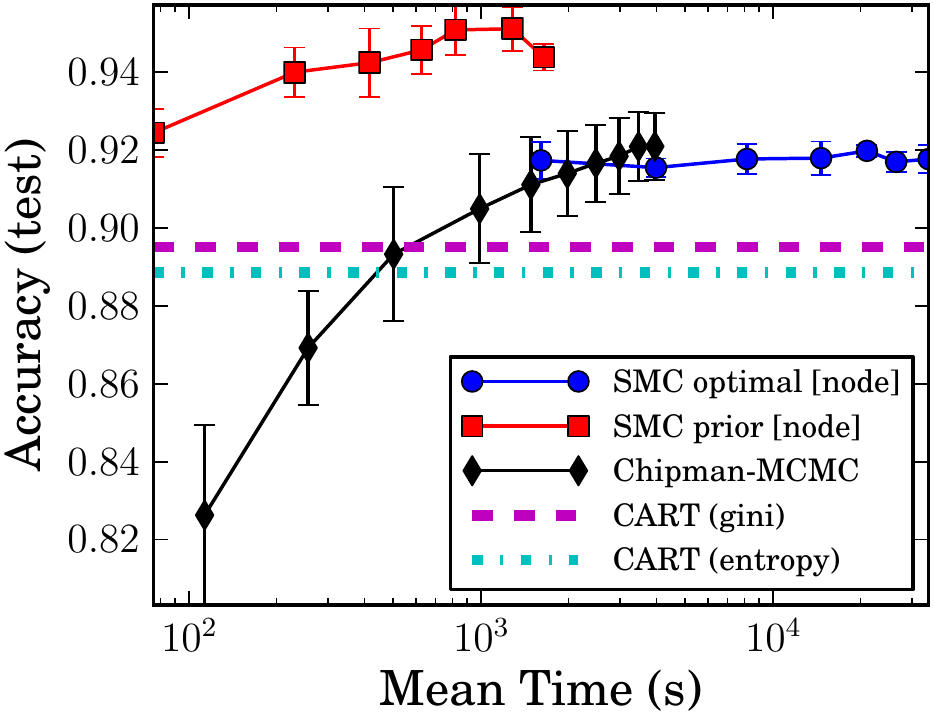}
        \end{subfigure}  
                        \begin{subfigure} 
                \centering
                \includegraphics[width=\sfigwidth]{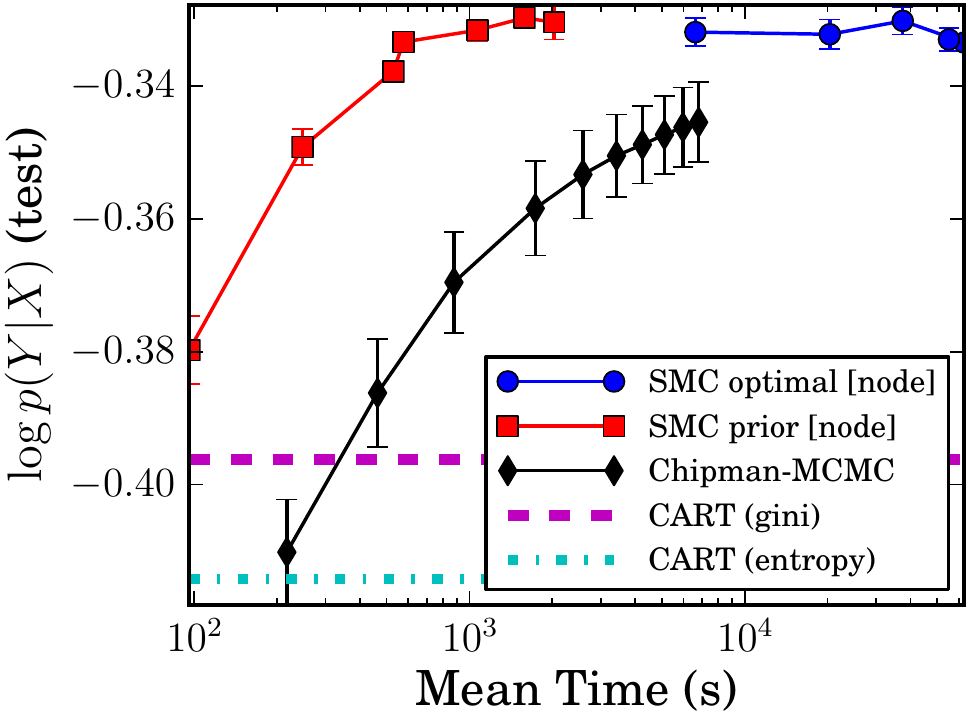}
        \end{subfigure}
                        \begin{subfigure} 
                \centering
                \includegraphics[width=\sfigwidth]{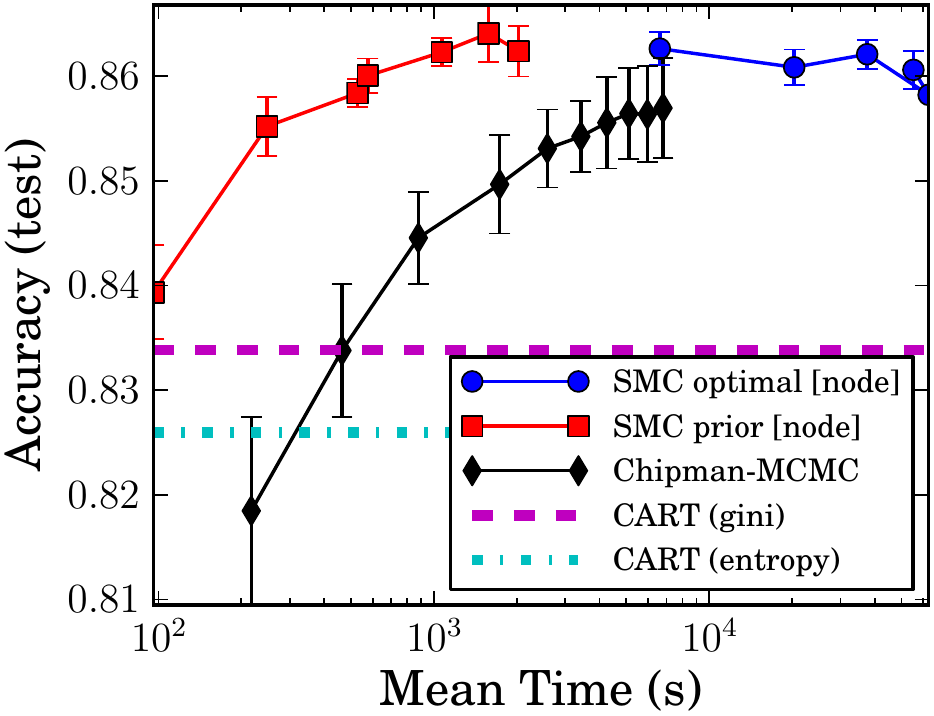}
        \end{subfigure}
        \caption{Hyperparameters: $\bm{\alpha=1.0}, {\alpha_s=0.95, \beta_s=0.5}$ (see main text for additional information).}
        \label{fig:logprobtest:acctest:time:alpha1}
\end{figure}

In this experiment, we evaluate the sensitivity of the runtime vs predictive performance comparison between SMC (\prior\ and \posterior\ proposals), MCMC and CART to the choice of hyper parameters $\alpha$ (Dirichlet concentration parameter) and $\alpha_s, \beta_s$ (tree priors).  
We consider only \nodewise\ expansion since it consistently outperformed \layerwise\ expansion in our previous experiments. In the first variant, we fix $\alpha=5.0$ (since we do not expect it to affect the timing results) and vary the hyper parameters from $\alpha_s=0.95,\beta_s=0.5$ to $\bm{\alpha_s=0.8,\beta_s=0.2}$ (bold reflects changes) and also consider intermediate configurations $\alpha_s=0.95,\bm{\beta_s=0.2}$ and $\bm{\alpha_s=0.8},\beta_s=0.5$. In the second variant, we fix
$\alpha_s=0.95,\beta_s=0.5$ and set $\bm{\alpha=1.0}$.  
Figures \ref{fig:logprobtest:acctest:time:2}, \ref{fig:logprobtest:acctest:time:3}, \ref{fig:logprobtest:acctest:time:4} and \ref{fig:logprobtest:acctest:time:alpha1} display the results on \pendigits\ (top row), and \magic\ (bottom row). The left column plots test $\log p(y|\bx)$ vs runtime, while the right column plots test accuracy vs runtime. The blue circles and red squares represent \posterior\ and \prior\ proposals respectively.
Comparing the results to Figure \ref{fig:logprobtest:acctest:time} (in main text), we observe that the trends are qualitatively similar to those observed for
$\alpha=5.0, \alpha_s=0.95,\beta_s=0.5$ in Section \ref{sec:expt:smcvsmcmc} (in main text): (i) SMC consistently offers a better runtime vs predictive performance tradeoff than MCMC, (ii) the \prior\ proposal offers a better runtime vs predictive performance tradeoff than the \posterior\ proposal, (iii) $\alpha=1.0$ leads to similar test accuracies as $\alpha=5.0$ (the predictive probabilities are obviously not comparable).

